\documentclass[pdflatex,sn-nature]{sn-jnl}

\usepackage{graphicx}%
\usepackage{multirow}%
\usepackage{amsmath,amssymb,amsfonts}%
\usepackage{amsthm}%
\usepackage{mathrsfs}%
\usepackage[title]{appendix}%
\usepackage{xcolor}%
\usepackage{textcomp}%
\usepackage{manyfoot}%
\usepackage{booktabs}%
\usepackage{algorithm}%
\usepackage{algorithmicx}%
\usepackage{algpseudocode}%
\usepackage{listings}%
\usepackage{longtable}
\usepackage{xurl}

\usepackage[group-separator={,},group-minimum-digits={3}]{siunitx}

\usepackage{lipsum} 

\theoremstyle{thmstyleone}%
\theoremstyle{thmstyletwo}%

\theoremstyle{thmstylethree}%

\newcommand\ours{ToothMCL}
\newcommand\ourdataset{\textbf{}CBCT-IOS3.8K}
\begin{document}
\newif\iffigures
\figurestrue
\title[Article Title]{Multimodal Contrastive Pretraining of CBCT and IOS for Enhanced Tooth Segmentation}


\author[1]{\fnm{Moo Hyun} \sur{Son}}\email{mhson@cse.ust.hk}
\author[1]{\fnm{Juyoung} \sur{Bae}}\email{jbaeaa@cse.ust.hk}
\author[1]{\fnm{Zelin} \sur{Qiu}}\email{zqiuak@cse.ust.hk}
\author[2]{\fnm{Jiale} \sur{Peng}}\email{u3012583@connect.hku.hk}
\author[3]{\fnm{Kai Xin} \sur{Li}}\email{497155504@qq.com}
\author[2]{\fnm{Yifan} \sur{Lin}}\email{yflin@hku.hk}
\author*[1,4,5,6,7]{\fnm{Hao} \sur{Chen}}\email{jhc@cse.ust.hk}

\affil[1]{\orgdiv{Department of Computer Science and Engineering}, \orgname{The Hong Kong University of Science and Technology}, \orgaddress{\state{Hong Kong SAR}, \country{China}}}

\affil[2]{\orgdiv{Division of Paediatric Dentistry and Orthodontics, Faculty of Dentistry}, \orgname{The University of Hong Kong}, \orgaddress{\state{Hong Kong SAR}, \country{China}}}

\affil[3]{\orgname{Delun Dental Hospital}, \orgaddress{Guangzhou}, \country{China}}

\affil[4]{\orgdiv{Department of Chemical and Biological Engineering}, \orgname{The Hong Kong University of Science and Technology}, \orgaddress{\state{Hong Kong SAR}, \country{China}}}

\affil[5]{\orgdiv{Division of Life Science}, \orgname{The Hong Kong University of Science and Technology}, \orgaddress{\state{Hong Kong SAR}, \country{China}}}

\affil[6]{\orgdiv{HKUST Shenzhen-Hong Kong Collaborative Innovation Research Institute}, \orgaddress{\state{Futian, Shenzen}, \country{China}}}

\affil[7]{\orgdiv{State Key Laboratory of Nervous System Disorders}, \orgname{The Hong Kong University of Science and Technology}, \orgaddress{\state{Hong Kong SAR}, \country{China}}}

\abstract{
Digital dentistry represents a transformative shift in modern dental practice. The foundational step in this transformation is the accurate digital representation of the patient's dentition, which is obtained from segmented Cone-Beam Computed Tomography (CBCT) and Intraoral Scans (IOS). Despite the growing interest in digital dental technologies, existing segmentation methodologies frequently lack rigorous validation and demonstrate limited performance and clinical applicability. This work pioneers the first multimodal pretraining framework for dental segmentation, addressing a critical and previously unmet challenge in the field. While prior approaches have predominantly relied on single-modality, ToothMCL establishes a new paradigm by integrating volumetric (CBCT) and surface-based (IOS) imaging through contrastive learning. By capturing modality-invariant representations through multimodal contrastive learning, our approach effectively models fine-grained anatomical features, enabling precise multi-class segmentation and accurate identification of Fédération Dentaire Internationale (FDI) tooth numbering. Along with the framework, we curated \ourdataset{}, the largest paired CBCT and IOS dataset to date, comprising 3,867 patients. We then evaluated \ours{} on a comprehensive collection of independent datasets, representing the largest and most diverse evaluation to date. Our method achieves state-of-the-art performance in both internal and external testing, with an increase of 12\% for CBCT segmentation and 8\% for IOS segmentation in the Dice Similarity Coefficient (DSC). Furthermore, \ours{} consistently surpasses existing approaches in tooth groups and demonstrates robust generalizability across varying imaging conditions and clinical scenarios. Our findings underscore the transformative potential of large-scale multimodal pretraining in digital dentistry and highlight the critical importance of effectively leveraging paired multimodal data. Our approach lays the foundation for enhanced clinical workflows, including caries detection, orthodontic simulation, and dental prosthesis design.
}


\maketitle

\section*{Introduction} 
Oral diseases remain one of the most pervasive global health issues, affecting over 3.5 billion individuals, which accounts for over 43\% of the global population as reported by the World Health Organization \cite{WHOOralHealth2022}. This widespread prevalence underscores the critical importance of dentistry, not only for clinical needs but also for enhancing the overall quality of life for a large portion of the global population. In modern dental practice, digital dentistry plays a crucial role in streamlining workflows and enhancing patient outcomes. Cone-Beam Computed Tomography (CBCT) visualizes 3D anatomical structures, including tooth morphology, alveolar bone, and surrounding tissues \cite{william2006clinicalCBCT}, while intraoral scans (IOS) provide high-resolution images of occlusal surfaces that are crucial for treatment planning and prosthesis design \cite{Mangano2017Intraoral}. However, these imaging modalities still require extensive manual and time-consuming analysis to identify and plan treatments \cite{Baldini2025}. Consequently, numerous research efforts now focus on automating key tasks such as caries detection \cite{Albano2024Caries, Adnan2024Caries, Negi2024Caries}, orthodontic treatment planning \cite{Liu2023Orthodontics, Deng2024TAPoseNet, Lei2024AutomaticTADPM}, and designing dental prostheses, including implants, crowns, and bridges \cite{Kong2024Prosthesis, Hosseinimanesh2023AICrown, Chau2024SingleMolar}. A fundamental step in these automated workflows is the accurate representation of patient dentition, achieved from the segmented model of both CBCT and IOS data. CBCT provides root morphology critical for implant planning and orthodontics, while IOS captures crown-level detail essential for prosthetic design and simulation.  Hence, accurate segmentation is a central requirement for advancing digital dentistry.

Recent developments in artificial intelligence (AI) and its integration into medical imaging have significantly advanced dental segmentation \cite{Chen2024Segmentation}. Convolutional neural networks (CNNs) \cite{lecun1998cnn} and multi-stage 3D U-Net-based approaches \cite{ronneberger2015unet, Cui2022FullyAutomaticAI} have demonstrated strong potential for segmenting CBCT images, enabling detailed analysis of oral anatomical structures. However, methods that require multiple stages can inadvertently propagate errors throughout the segmentation process. Transformer-based architectures \cite{vaswani2017attention} such as AMASS \cite{Gillot2022AMASS} and Mamba architectures \cite{Gu2023mamba} like T-Mamba \cite{hao2024tmamba} have primarily excelled in binary or instance segmentation tasks. Binary segmentation, although efficient for general positioning, lacks the precision needed to distinguish individual teeth for specific clinical tasks such as orthodontic planning or tooth morphology monitoring. Meanwhile, instance segmentation, despite providing more detail, often assigns arbitrary instance IDs that do not conform to established clinical standards, limiting their applicability.

IOS segmentation methodologies leveraging graph neural networks (GNNs) \cite{scarselli2009gnn} have shown significant promise. GNN-based approaches utilize hierarchical edge convolution to integrate rich topological information, allowing for context-aware segmentation, which consistently outperforms traditional clustering-based methods. Transformer-inspired, point-based techniques such as TGNet \cite{BenHamadou2022} have further enhanced accuracy and topological sensitivity compared to conventional k-means centroid clustering. Despite these improvements, conventional centroid-based clustering remains challenged by oversimplified spherical assumptions, reducing accuracy at tooth boundaries, particularly in densely arranged teeth.

Despite these advancements, existing methodologies exhibit several significant limitations restricting their clinical effectiveness. Binary segmentation methods lack sufficient detail for precise tooth identification, essential for clinical tasks such as orthodontic planning or tooth root morphology analysis. Instance segmentation approaches, although more detailed, frequently face semantic consistency issues, with arbitrarily assigned identifiers that fail to comply with clinical standards like the Federation Dentaire Internationale (FDI) numbering system \cite{ISO3950_2016}. Additionally, validation efforts often utilize small, limited datasets, which restrict the generalizability and clinical robustness of these methods \cite{Schwendicke2020Artificial}. Lastly, many existing solutions attempt to enhance performance by incorporating overly complex submodules, inadvertently introducing noise and compromising the model’s understanding of intricate oral anatomical features. Addressing these limitations necessitates the development of innovative multi-class segmentation frameworks capable of precise tooth identification in compliance with clinical standards, supported by comprehensive validation across diverse clinical datasets.

\iffigures
\begin{figure}[!]
    \centering
    \includegraphics[width=\textwidth]{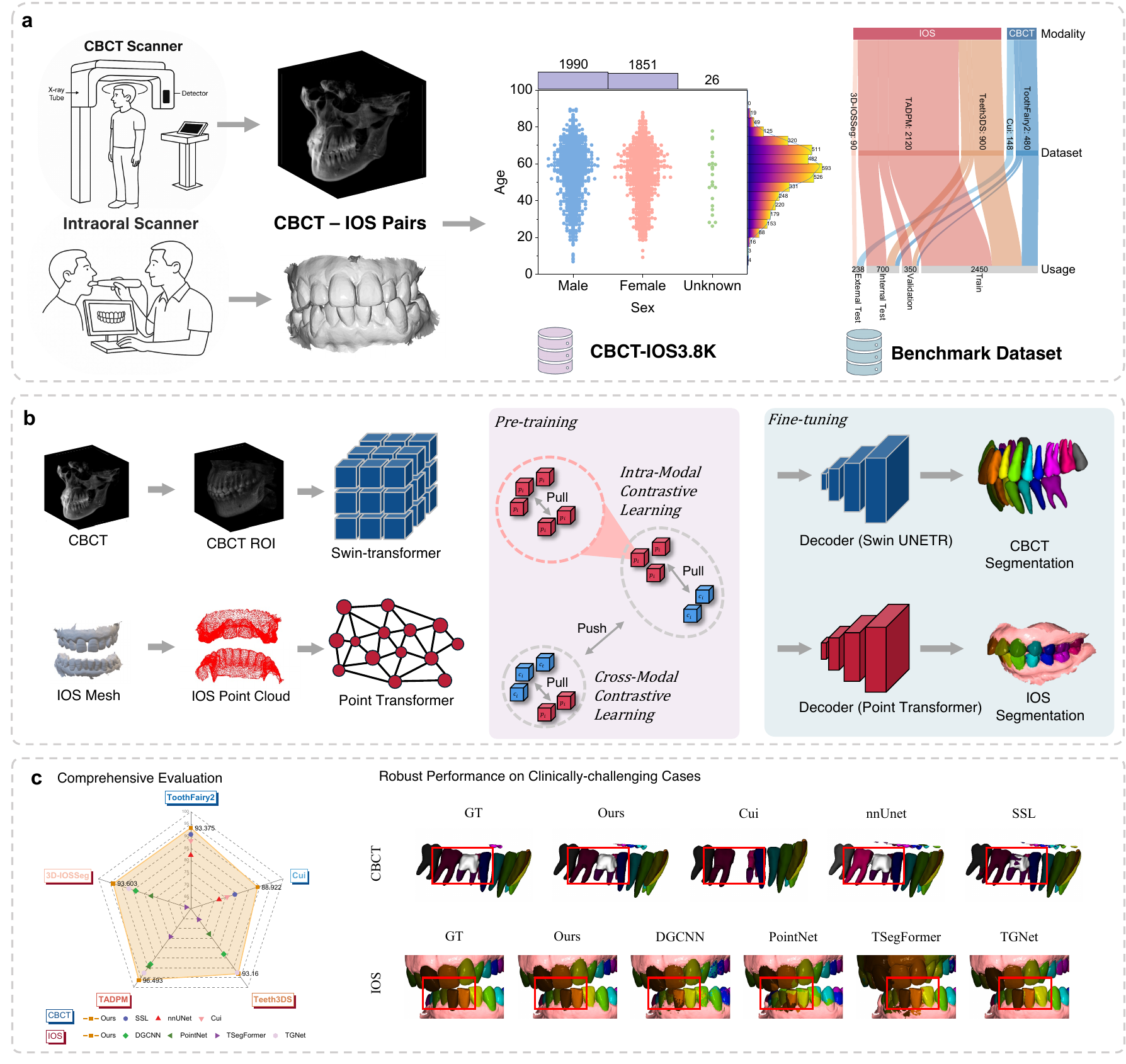}
    \caption{\textbf{Overview of our proposed \ourdataset{} dataset and \ours{} framework.} \textbf{a}. The largest CBCT-IOS paired dataset \ourdataset{} was collected. \textbf{b}. The proposed Tooth Multimodal Contrastive Learning for pretraining approach leverages paired CBCT and IOS data to learn unified representations across CBCT and IOS data. \textbf{c}. Extensive validation of the proposed method in the largest and most comprehensive public datasets.}
    \label{fig:overview}
\end{figure}
\fi

In recent years, there has been growing interest in multimodal contrastive learning \cite{chen2020Contrastive, radford2021Clip}. A multimodal contrastive learning framework aims to align the representations of different data modalities, such as volumetric CBCT and surface-based IOS, in a common latent space. In dentistry, CBCT and IOS modalities provide complementary information; where CBCT provides volumetric context about the internal anatomy, IOS captures high-resolution surface morphology. If common anatomical information is learned through alignment using contrastive learning objectives, multimodal models can learn more robust and discriminative representations, improving the accuracy of segmentation and generalization across imaging modalities. The multimodal aspect of the fusion framework is especially promising to promote better outcomes on clinical tasks, such as accurately localizing a tooth and determining appropriate treatment plans.

On this theoretical foundation, we present \ours{}, the first large-scale multimodal pretraining framework for dental imaging. \ours{} leverages the largest paired CBCT-IOS dataset created to date and jointly encodes volumetric and surface features into a common representation space. Utilizing both modalities in a common space with our framework allows for cross-modal knowledge transfer and strengthens the segmentation of multi-class teeth in various imaging conditions. Our framework demonstrated improved generalization capabilities across patient anatomies and scanner protocols for representing the data and forming robust clinical pipelines for wider use. 
\newline
\space{}
\newline \noindent Our primary contributions can be summarized as:
\begin{enumerate}
\item \textbf{\ourdataset{}: The Largest CBCT-IOS Paired Dental Dataset}.
We introduce \ourdataset{}, the largest paired dental imaging dataset to date, comprising CBCT and IOS data from over 3,867 patients. This extensive dataset captures diverse morphological variations, clinical conditions, and imaging protocols, laying a robust foundation for dental AI research.

\item \textbf{\ours{}: Tooth Multimodal Contrastive Learning for Pretraining}.
We propose \textbf{\ours{}}, a novel dental pretraining framework that utilizes advanced multimodal contrastive learning to unify volumetric (CBCT) and surface-based (IOS) dental imaging modalities. \ours{} effectively aligns fine-grained morphological features across both modalities, significantly improving feature representation and anatomical precision.

\item \textbf{Extensive Public Benchmark Validation and State-of-the-Art Performance}.
We rigorously validate \ours{}'s performance using the largest and most comprehensive public CBCT and IOS dataset, featuring diverse clinical sources. Our experimental evaluations demonstrate that \ours{} achieves state-of-the-art performance, surpassing existing approaches in 12\% in CBCT and 8\% in IOS segmentation Dice Similarity
Coefficient (DSC) and robustness to clinical variability. These findings establish \ours{} as a new foundation for digital dentistry with immediate applicability for enhancing clinical outcomes and workflows.
\end{enumerate}

\section*{Results} 
\subsection*{Overview of Proposed \ours{}}
We structured our workflow into three stages: pretraining, fine-tuning, and evaluation. In the pretraining stage, we employed multimodal contrastive learning using our \ourdataset{} dataset to capture modality-invariant representations between CBCT and IOS modalities. Although each modality encodes oral anatomy differently (voxel-based and surface-based), both represent the same underlying dental structures. A large-scale paired dataset at this stage is crucial for aligning these representations, effectively learning transferable features, and enhancing cross-modality information sharing. During fine-tuning, we applied fully supervised learning individually to modality-specific encoders on public datasets to further refine segmentation accuracy. Finally, in the evaluation stage, we assessed the robustness and adaptability of our model by testing on both internal and external datasets.

\subsection*{Overview of \ourdataset{}}
Since the performance of pretraining highly relies on large-scale data, we curated the largest CBCT-IOS paired dataset, the \ourdataset{}. As outlined in Extended Data Table~\ref{table:datasets}, this dataset consists of 3,867 paired CBCT and IOS scans acquired under standardized imaging protocols. These scans underwent basic quality control checks before being designated for pretraining. All of these scans remain unlabeled, as the primary objective of this phase is to learn modality-invariant representations in a self-supervised manner.

\subsection*{Evaluation}
Following fine-tuning, we evaluated \ours{} across diverse datasets. Our evaluation assessed the model's accuracy, generalizability, and consistency in delineating tooth structures across both CBCT and IOS modalities, with particular attention to performance in challenging anatomical regions and on out-of-distribution data.

We evaluated the performance of our method using the DSC, which quantifies the overlap between predicted and ground truth segmentations. The DSC is defined as:
\begin{equation}
\text{DSC} = \frac{2|A \cap B|}{|A| + |B|},
\end{equation}
where $A$ and $B$ represent the sets of predicted and ground truth voxels, respectively. The evaluation was also conducted by grouping teeth into each category: central incisors, lateral incisors, canines, premolars (first, second, and third), and molars (first, second, and third), assessing performance individually for each tooth group to provide detailed insights into model accuracy across different anatomical regions.
\iffigures
\begin{figure}[t]
    \centering
    \includegraphics[width=\textwidth]{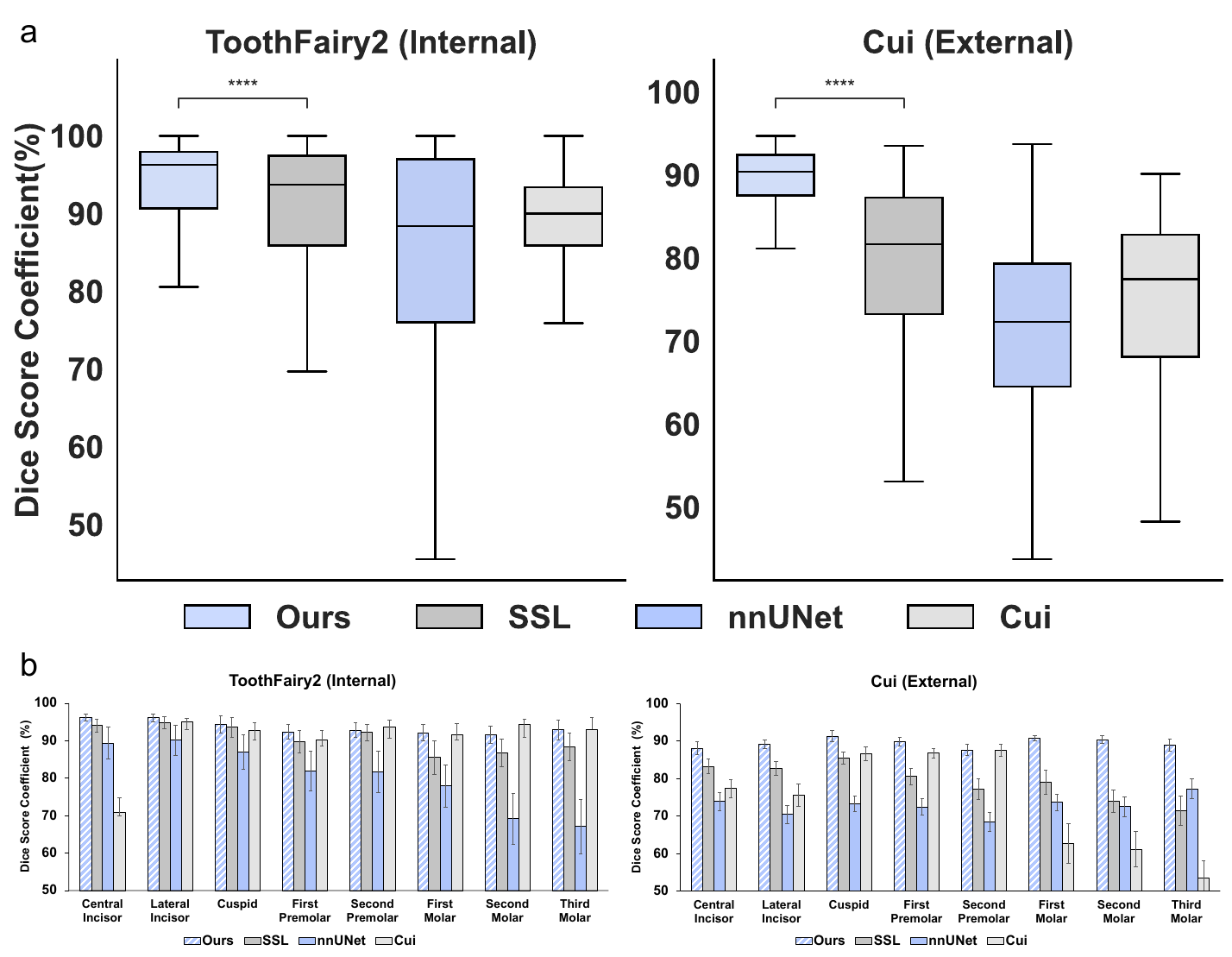}
    \caption{\textbf{a}. Box plot comparing DSC (\%) of competing methods for CBCT tooth segmentation on ToothFairy2 (Internal) and Cui (External) datasets. Statistical significance is indicated by asterisks (****: p $<$ 0.0001). \textbf{b}. Per-tooth group CBCT segmentation results on the ToothFairy2 (Internal) and Cui (External) datasets, measured using the DSC (\%) with a 95\% Confidence Interval (CI).}
    \label{fig:cbct_overall}
\end{figure}

\begin{figure}[!htbp]
    \centering
    \includegraphics[width=\textwidth]{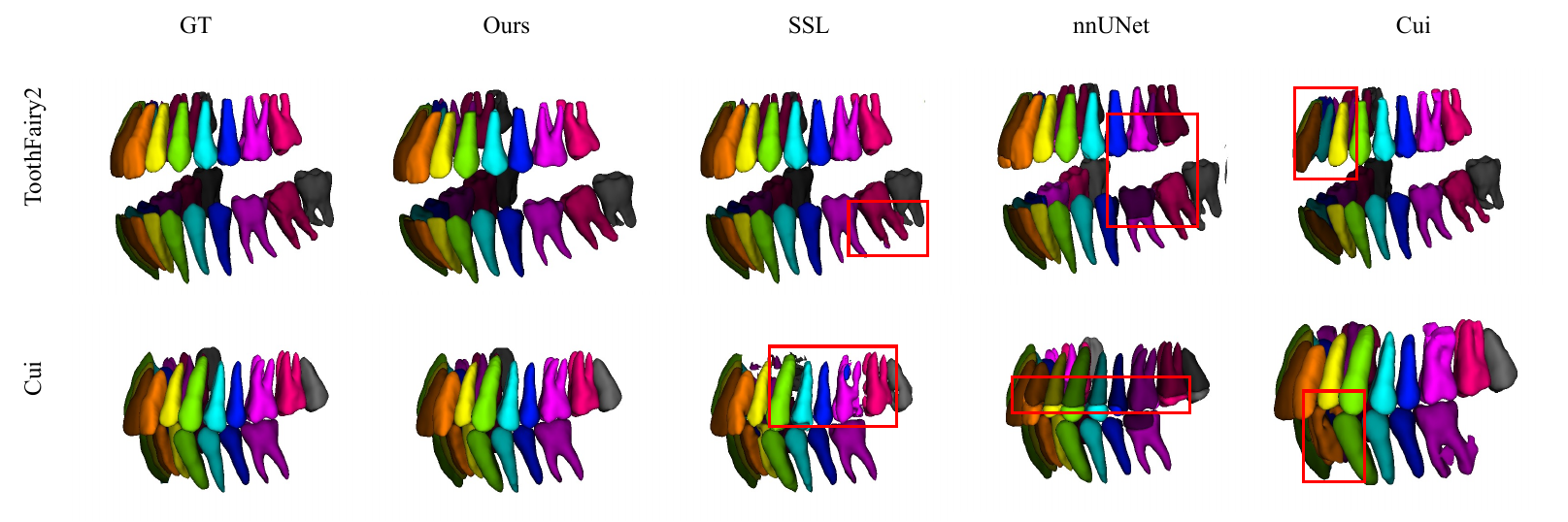}
    \caption{Qualitative evaluation of CBCT tooth segmentation results, visual comparisons between ground truth and predicted segmentations. Different colors represent individual tooth instances. Red boxes indicate areas with severe defects in the segmentation results.}
    \label{fig:cbct_vis}
\end{figure}
\fi

\subsection*{CBCT Segmentation} 
For CBCT segmentation, we compare our approach with SSL \cite{Tang2022SelfSupervised}, Cui \cite{Cui2022FullyAutomaticAI}, and nnUNet \cite{Isensee2021nnUNet}, each represents a different paradigm in medical image segmentation. SSL \cite{Tang2022SelfSupervised} employs self-supervised learning, leveraging unlabeled data to pretrain models for enhanced feature extraction, making it a relevant baseline for evaluating the benefits of our pretraining approach. Cui \cite{Cui2022FullyAutomaticAI} proposes a fully automatic AI-based CBCT segmentation method, which serves as a strong domain-specific benchmark due to its specialized focus on dental structures. nnUNet \cite{Isensee2021nnUNet}, a state-of-the-art automated deep learning framework, is widely regarded as a robust and adaptable baseline for medical image segmentation, offering a fair comparison against a highly optimized, general-purpose model. By including these three methods, we assess the effectiveness of our approach across self-supervised learning, domain-specific methodologies, and general deep learning architectures, ensuring a comprehensive evaluation.

\subsubsection*{Evaluation on Internal Dataset}
Our proposed method significantly outperformed all baseline models, achieving state-of-the-art performance on the internal testing sets. Specifically, it achieved a DSC of 93.38\% (95\% CI: 92.02–94.73), which signifies the top overall score with a statistically significant improvement (p $<$ 0.0001) compared to other methods.

A detailed analysis across different tooth groups further highlights the consistency of our model’s performance. Anterior teeth (Central Incisor, Lateral Incisor, and Canine) consistently exhibited superior segmentation accuracy, with DSC of 96.21\% (95\% CI: 95.34–97.08), 96.10\% (95\% CI: 95.03–97.17), and 94.37\% (95\% CI: 92.01–96.74), respectively. In contrast, other methods showed greater variability among these groups, often struggling to maintain such consistently high scores.

The molar regions, particularly the second and third molars, present unique challenges due to structural variability among patients, which may be complicated by the anatomical absence of these teeth in certain individuals. Our model achieved strong performances in DSCs of 91.56\% (95\% CI: 89.31–93.80) and 92.92\% (95\% CI: 90.28–95.55). Importantly, all tooth groups in our approach exceeded a 90\% DSC threshold. This superior performance derives from the model’s ability to capture and adapt to the inherent structural variability of different tooth groups—especially the frequently underrepresented third molars, which enables highly reliable segmentation even under complex anatomical conditions.

These results establish our method as the most reliable and consistent performer among diverse tooth groups, reinforcing its utility in real-world dental applications

\subsubsection*{Evaluation on External Dataset}
Significant variability exists across CBCT datasets due to differences in scanner devices, voxel resolutions, and patient populations. Under the proposed framework, the pretrained model extracted robust and domain-invariant features, enhancing generalizability. To validate this, we evaluated our method on an out-of-distribution CBCT image from Cui \cite{Cui2022FullyAutomaticAI}.

On the external test set, our model achieved an overall DSC of 88.92\% (95\% CI: 87.97–89.87), demonstrating strong generalization to unseen clinical data (Figure \ref{fig:cbct_overall}). Notably, our approach outperformed the next best-performing model (Cui \cite{Cui2022FullyAutomaticAI}) by a substantial 12\% margin, which achieved 79.07\% (95\% CI: 77.27–80.87). Furthermore, our method exhibited a significantly smaller confidence interval, highlighting its stability even in external datasets. The performance difference between our method and the next best-performing model was statistically significant (p $<$ 0.0001), confirming the superiority of our approach.

Our model maintained the best performance across all tooth groups, including the third molar, where conventional models typically experience a notable drop in accuracy. This reinforces the model’s ability to handle underrepresented and clinically challenging cases, further solidifying its applicability in real-world dental workflows. 

Figure~\ref{fig:cbct_vis} illustrates a comparative analysis of segmentation outputs, highlighting our model’s ability to maintain anatomical fidelity. A common segmentation artifact arises when the model fails to accurately detect all teeth, leading to incomplete segmentation outputs, or incorrectly identifies certain regions despite an otherwise accurate segmentation. These errors are particularly evident in Cui \cite{Cui2022FullyAutomaticAI} and nnUNet \cite{Isensee2021nnUNet}, as shown in Figure~\ref{fig:cbct_vis}, where their models frequently fail to consistently delineate complete tooth structures. In contrast, our approach significantly mitigates these artifacts, ensuring more complete and anatomically precise segmentations. This improvement is primarily attributed to our multimodal contrastive pretraining, which effectively learns robust, modality-invariant features, enhancing the model's ability to generalize across varied anatomical structures.

\subsection*{IOS Segmentation} 
PointNet \cite{qi2017pointnet}, DGCNN \cite{wang2019dgcnn}, TSegFormer \cite{lee2023tsegformer}, and TGNet \cite{BenHamadou2022} represent key advances in point cloud processing and segmentation, making them strong benchmarks for comparing IOS segmentation performance. PointNet \cite{qi2017pointnet} pioneered a direct approach to processing raw point clouds without requiring voxelization or other preprocessing steps, making it a foundational method in point-based 3D segmentation. DGCNN \cite{wang2019dgcnn} leverages dynamic graph construction, which is effective in representing irregular structures like teeth and maintaining robust performance across varying resolutions and scan qualities. TSegFormer \cite{lee2023tsegformer}, on the other hand, combines the transformer paradigm with multi-scale feature extraction, providing a well-rounded approach to segmenting intricate dental structures. TGNet \cite{BenHamadou2022} focuses on accurate boundary point sampling by integrating a secondary network built on point transformers, which enhances tooth instance separation and enables reliable grouping of closely located teeth. By including these methods, we aim to evaluate our approach against models that span foundational point cloud processing architectures, multi-scale attention mechanisms, dental-specific architectures, and graph-based techniques. This diverse set of baselines ensures a comprehensive assessment of our model’s relative strengths, robustness, and applicability in IOS segmentation tasks.

\subsubsection*{Evaluation on Internal Dataset}
On the internal testing sets, our proposed method demonstrated significant improvements overall baseline models, achieving state-of-the-art results. Specifically, it achieved overall DSC of 93.16\% (95\% CI: 92.58–93.74) and 96.49\% (95\% CI: 96.26–96.73) for the Teeth3DS \cite{BenHamadou2022} and TADPM \cite{Lei2024AutomaticTADPM} datasets, respectively. These results not only represent the highest overall scores but also indicate statistically significant (p $<$ 0.05, p $<$ 0.0001) gains compared to existing approaches.

When examining performance across different tooth groups, our approach proved to be the most consistent. Anterior teeth (Central Incisor, Lateral Incisor, and Cuspid) achieved DSCs of 93.61\% (95\% CI: 92.79–94.44), 93.42\% (95\% CI: 92.67–94.17), and 92.71\% (95\% CI: 91.68–93.73), respectively in Teeth3DS \cite{BenHamadou2022}. This high level of accuracy was maintained even in the structurally variable molar regions (Second Molar and Third Molar), with DSC of 97.59\% (95\% CI: 97.25–97.93) and 91.68\% (95\% CI: 90.31–93.04) in TADPM \cite{Lei2024AutomaticTADPM}. Notably, every tooth group exceeded a 90\% DSC threshold—a level of consistency not observed in other methods, which often exhibit significant fluctuations. By consistently performing well across all tooth groups, including the underrepresented third molar, our model sets a new standard for internal segmentation performance.

\subsubsection*{Evaluation on External Dataset}
To evaluate the generalizability of our approach, we tested on the out-of-distribution external IOS dataset, 3D-IOSSeg. External datasets often introduce significant variability due to differences in scanner configurations, software, and resolution. Despite these challenges, our model maintained strong performance, achieving an overall DSC of 93.60 (95\% CI: 93.05–94.16). This represents a substantial 7\% improvement over the next best-performing method (TGNet \cite{BenHamadou2022}) with statistical significance (p $<$ 0.0001), which attained a DSC of 86.86 (95\% CI: 84.92–88.80). Furthermore, our approach exhibited a smaller confidence interval, underscoring its stability and reliability across diverse real-world clinical scenarios.

When analyzing individual tooth groups, our model consistently achieved the best performance, except in the third molar group. Overall, these results confirm the model’s robustness and its capacity to handle underrepresented and clinically challenging cases.

Figure~\ref{fig:ios_vis} illustrates a comparative analysis of segmentation outputs, highlighting the superior performance of our model. Common segmentation artifacts, such as missing teeth or misidentified regions, were significantly reduced. In contrast, existing methods, such as PointNet \cite{qi2017pointnet} and DGCNN \cite{wang2019dgcnn}, frequently failed to delineate clear boundaries, as evident in Figure~\ref{fig:ios_vis}. By mitigating these artifacts, our model consistently produces more anatomically accurate and complete segmentations, ensuring greater reliability in clinical workflows.

\iffigures
\begin{figure}[!t]
    \centering
    \includegraphics[width=\textwidth]{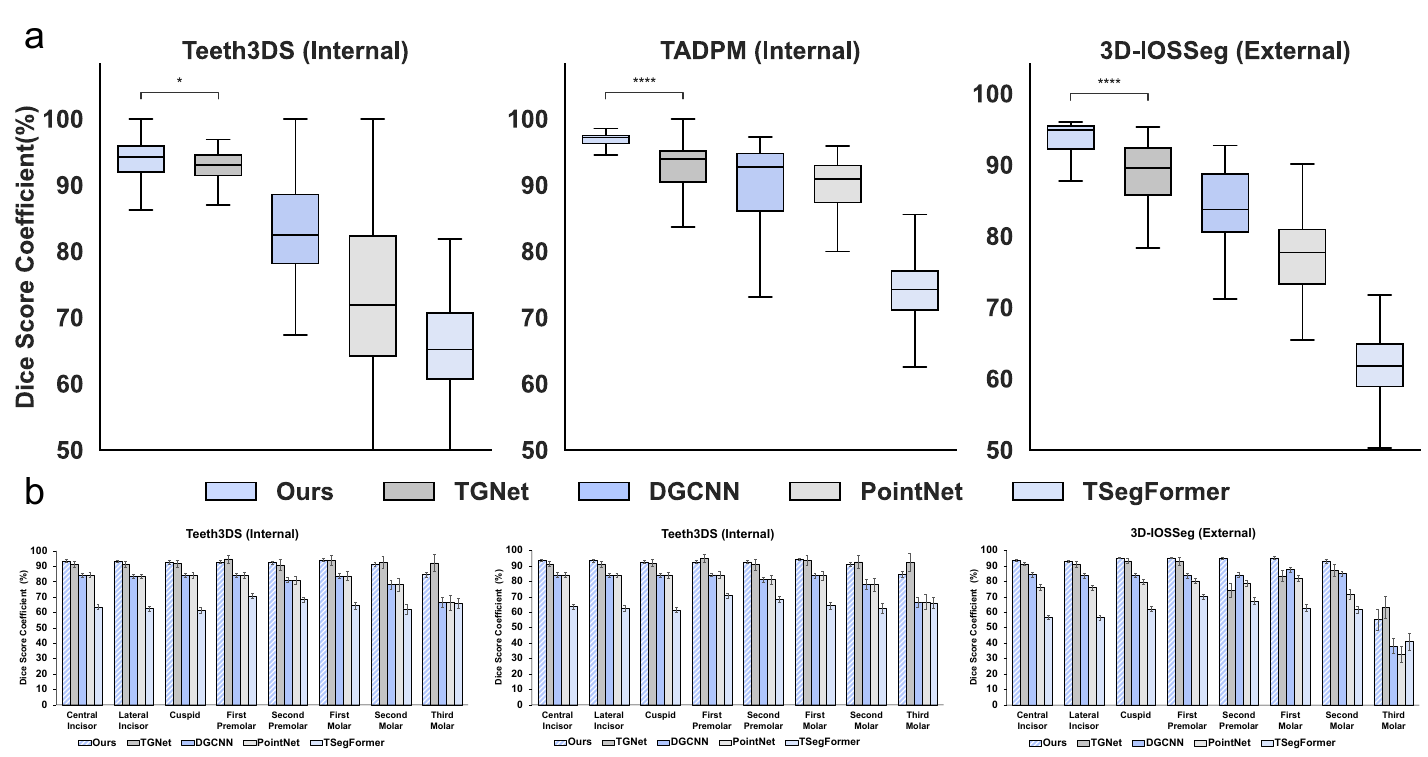}
    \caption{\textbf{a}. Box plot comparing the DSC (\%) of competing deep learning methods for IOS tooth segmentation. Results are shown for all Teeth3DS (Internal), TADPM (Internal), and 3D-IOSSeg (External) datasets. Statistical significance is indicated by asterisks (*: p $<$ 0.05, ****: p $<$ 0.0001). \textbf{b}. Per-tooth group IOS segmentation results on the Teeth3DS (Internal), TADPM (Internal), and 3D-IOSSeg (External) datasets, measured using the DSC (\%) with a 95\% Confidence Interval (CI).}
    \label{fig:ios_overall}
\end{figure}

\begin{figure}[!t]
    \centering
    \includegraphics[width=\textwidth]{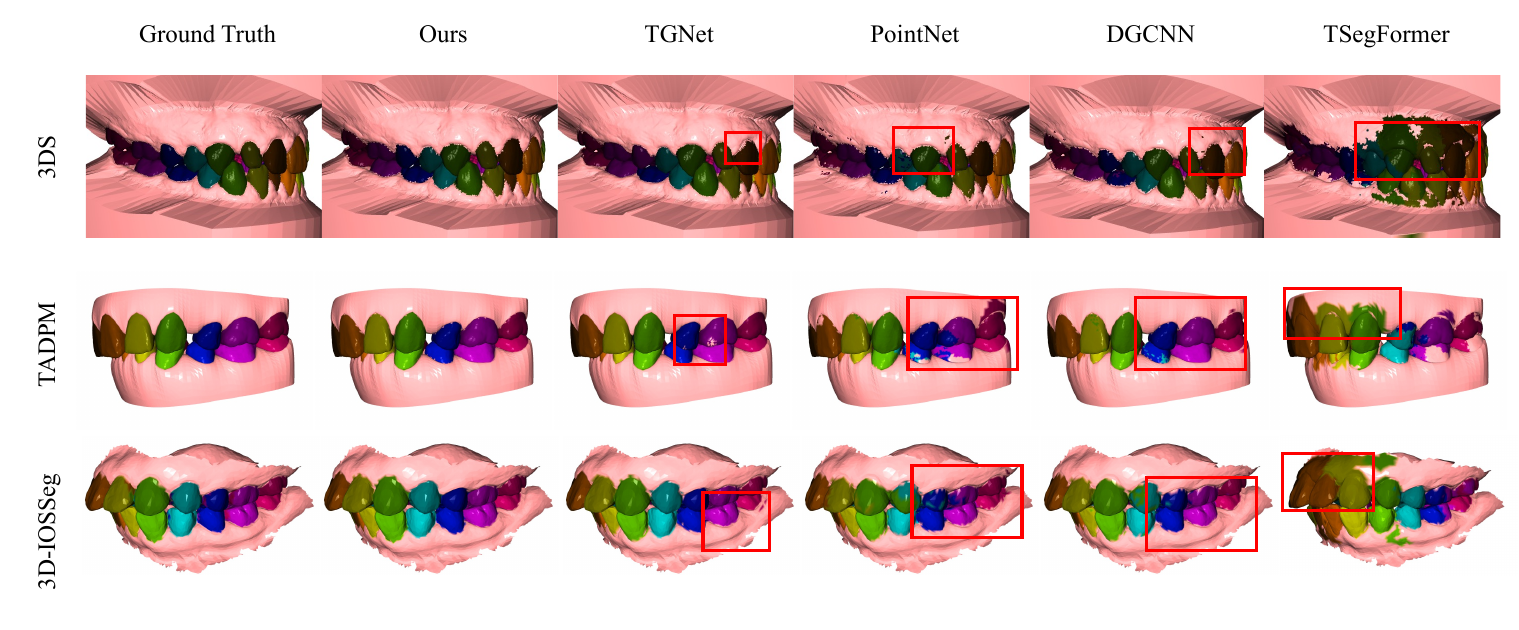}
    \caption{Qualitative evaluation of IOS tooth segmentation results, visual comparisons between ground truth and predicted segmentations. Different colors represent individual tooth instances. Red Boxes indicate areas with severe defects in the segmentation results.}
    \label{fig:ios_vis}
\end{figure}

\begin{figure}[!t]
    \centering
    \includegraphics[width=\textwidth]{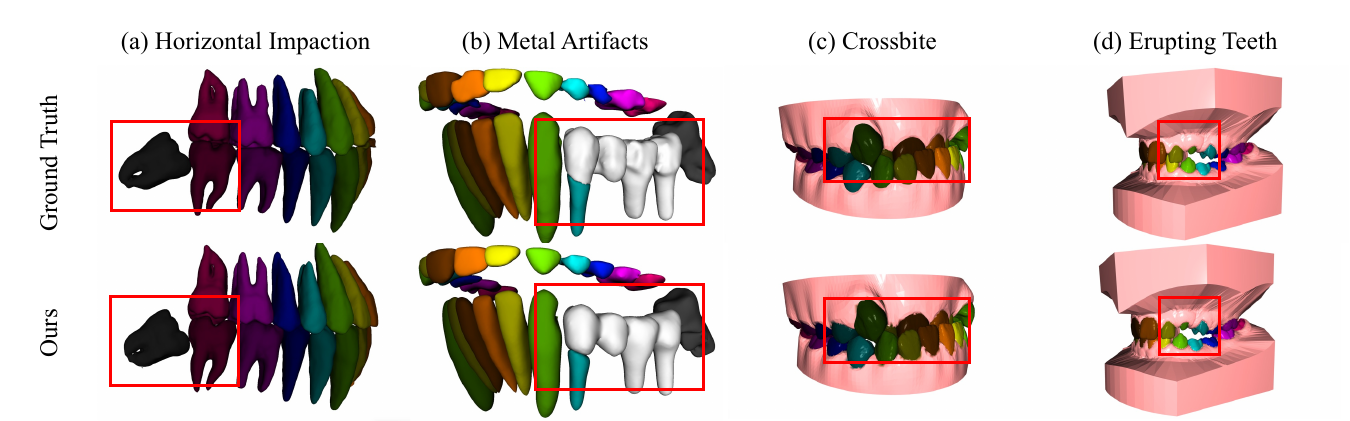}
    \caption{Visualization of segmentation results on clinically challenging cases, including (a) horizontally impacted third molar, (b) presence of metal-induced artifacts, (c) skeletal Class III malocclusion (crossbite), and (d) impacted teeth. Each example illustrates the model’s robustness under adverse anatomical and imaging conditions.}
    \label{fig:clinical_hard}
\end{figure}
\fi

\subsection*{Evaluation on Clinically Challenging Cases}
Clinically challenging cases such as teeth in atypical orientations, partial eruptions, significant malocclusions, and pronounced imaging artifacts from dental restorations often hinder accurate dental segmentation. For instance, horizontally impacted molars lodged sideways in the jaw present significant segmentation challenges, as conventional imaging methods frequently fail by either missing these teeth entirely or incorrectly merging them with surrounding bone. Similarly, partial visibility of crowns or teeth significantly overlapping due to extreme malocclusions, such as an exaggerated crossbite, complicates accurate delineation. Moreover, dental appliances like metal crowns, implants, or braces create strong artifacts, producing streaks and shadows that traditionally lead to misidentification or omission of dental structures.

\ours{} shows robust performance in these challenging clinical scenarios. As seen in Fig.~\ref{fig:clinical_hard} (a), a molar that is impacted in the alveolar bone is accurately segmented. Reliability in the segmentation of teeth is particularly important for processes like the extraction of horizontally impacted wisdom teeth, which the clinician needs to avoid damaging the patient's nerve and ensure the tooth is entirely removed. \ours{} also helps solve imaging distortions from metal restorations, including crowns, bridges, and implants, as seen in Fig.~\ref{fig:clinical_hard} (b). Additionally, in Fig.~\ref{fig:clinical_hard} (c), \ours{} demonstrates accurate separation of lower and upper incisors that are exceptionally overlapped, ultimately benefiting the exactness of the planning process of orthodontic treatment, as the model correctly established the shape and position of teeth. The clinical utility of \ours{} is supported by the robust performance against these edge cases and by reducing manual adjustments for misidentifications that might result in complications or delays in treatment. Furthermore, as shown in Fig.~\ref{fig:clinical_hard} (d), the model successfully segments an erupting canine-a challenging case commonly seen during the mixed dentition period. This capability is particularly valuable for pediatric diagnosis and treatment planning, where accurate identification of erupting teeth is crucial. High accuracy of \ours{} in diverse anatomical conditions improves patient outcomes by reducing surgical complications and providing a better degree of accurate, effective treatment plans. By leveraging multimodal contrastive pretraining on \ourdataset{}, which includes a wide range of anatomical variations and clinical conditions—such as impacted teeth, malocclusions, and metal artifacts—\ours{} gains a deeper understanding of oral anatomy. This enables it to generalize more effectively across diverse patient presentations, reinforcing its clinical relevance and demonstrating the robustness of the proposed learning framework.

\subsection*{Ablation Study}
In Fig.~\ref{fig:ablation}, we conduct an analysis of two ablation studies to quantify the precise impact of the initialization strategy and pretraining scale. Across all five benchmarks, pretraining improves overall DSC by 2.7–9.9 percentage points (pp) (Extended Data Table \ref{table:ablation_results}). The performance gain is smallest on ToothFairy2 \cite{2024TMI}, where in-domain images already resemble the finetuning data yet still reach a statistically significant 2.69 pp improvement (p $<$ 0.0001).  On the out-of-distribution Cui \cite{Cui2022FullyAutomaticAI} dataset, the improvement is greatest by 9.85 pp, underscoring that the encoder internalization data source invariant tooth morphology. Standard error decreases in every case, indicating not just higher accuracy but markedly more stable predictions.

We subsampled \ourdataset{} at 25\%, 50\%, 75\%, and 100\% of its full size. Mean DSC rises monotonically with each increment: +3.79 pp from 25\% to 50\% and +5.87 pp from 75\% to 100\%. Importantly, the slope of the curve does not indicate any sign of diminishing returns—no saturation is observed even after 3,867 paired scans, suggesting that further paired data would likely still yield significant improvements.

Altogether, the findings indicate that multimodal pretraining improves the latent space of the encoder to be more knowledgeable about anatomical structures. Pretraining on a heterogeneous dataset also allows the encoder to identify unusual morphological variations, as well as scanner artifacts. It is also important to recognize that there is no upper limit on performance, so paths to future gains in performance could include enlarging our dataset or augmenting it with synthetic but anatomically valid pairs.

\iffigures
\begin{figure}[!t]
    \centering
    \includegraphics[width=\textwidth]{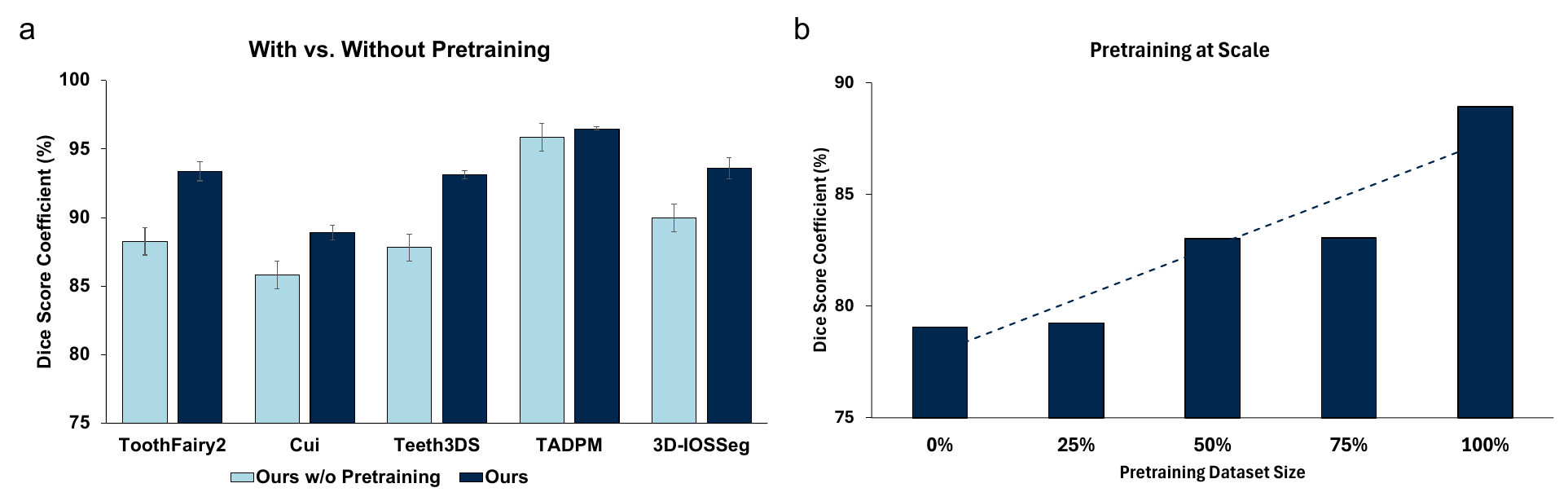}
    \caption{\textbf{a}. Comparison of overall DSC (\%) between models trained with and without pretraining across five datasets. Bars denote the mean DSC with standard error. \textbf{b}. Segmentation performance improves proportionally with the size of the pretraining dataset.}
    \label{fig:ablation}
\end{figure}
\fi
\section*{Discussion} 
This research presents \ours{}, a novel multimodal contrastive pretraining framework designed to enhance tooth segmentation and identification by merging volumetric CBCT data and surface-based IOS modality. Our findings reveal that multimodal contrastive learning substantially enhances model performance across all clinical scenarios with state-of-the-art accuracy based on internal and external validations. These enhancements represent significant advances toward resolving the clinical challenges associated with accurate, efficient, and automated tooth segmentation. 

One significant contribution of our framework is its ability to robustly address clinically challenging scenarios, such as horizontally impacted teeth, artifacts due to metal, and various malocclusion, such as a skeletal Class III crossbite. By explicitly training on modality invariant representations based on the largest available paired CBCT-IOS dataset, we demonstrate not only enhanced generalizability but also robustness, which is necessary for adoption into clinics, especially with the potential for same-day workflows - a significant time saver for orthodontic and prosthetic treatments.

Comparatively, \ours{} advances beyond previous single-modality approaches, such as the work by Cui \cite{Cui2022FullyAutomaticAI} and TGNet \cite{BenHamadou2022}, by explicitly leveraging the complementary strengths of CBCT and IOS data. Where prior methods showed limitations either in capturing fine anatomical details or in generalizing across varied clinical imaging conditions, our multimodal approach successfully addresses these gaps. Our results, achieving DSC improvements of approximately 12\% over previous benchmarks, underscore the critical importance of multimodal data utilization in digital dentistry.

Furthermore, \ours{} demonstrates potential as a foundational model for broader dental applications beyond segmentation alone. The unified latent space representations learned during multimodal pretraining offer significant promise for extension into tasks such as disease classification, restorative planning, and orthodontic simulations. As dental care shifts to a multidisciplinary digital approach, our framework could also serve as a core integration platform for these broader clinical workflows. Though there are many benefits, the results of our work have also pointed to challenges and areas for further improvement and research. 

\ours{} has undergone the most extensive validation in AI-driven dental segmentation, making it the largest benchmark study in the field. Unlike prior works that rely on single data-source validation, \ours{} has been rigorously tested on multiple independent datasets, ensuring robustness. However, while it has been validated on the largest benchmark to date, the dental field still lacks diverse validation datasets. Performance may still be suboptimal for rare dental conditions or imaging artifacts that are not well represented in current datasets. Expanding validation efforts to include underrepresented geographic regions and a broader spectrum of clinical imaging variations would further confirm its applicability.

Beyond segmentation, \ours{}’s scalable and adaptable architecture allows for potential applications in digital dentistry. The pretrained encoder could be used for varying applications beyond tooth segmentations, such as disease classification, restorative planning, and orthodontic treatment simulations, allowing for it to be viewed as a foundation model for dental research. The pretrained encoder may be directly applicable or fine-tuned or by applying and adapting to a specific application that employs labeled datasets. Further research could also explore options in transfer learning and multimodal foundation models for use for a seamless transfer to dental applications without full retraining of a model.

Although \ours{} is specifically tuned for CBCT and IOS data with unprecedented segmentation accuracy and reliable feature extraction, it cannot accommodate other common imaging techniques used in dentistry, such as panoramic and cephalometric x-rays, which are important in general dentistry and orthodontics. Furthermore, the adaptation of new imaging modalities can be less straightforward due to variations in spatial resolution, the distribution of features among subjects, and varying standards of clinical annotation. One way to improve the adaptability of \ours{} to a wider range of dental indications is by conducting multimodal pretraining of new modalities, including panoramic and cephalometric x-rays while devising a hybrid fusion model that extracts shared anatomical features or performs the same segmentation task with other modalities without full retraining.

In conclusion, \ours{} marks a step forward in the field of dental imaging while improving multi-class segmentation accuracy, tooth identification, and multimodal learning. Addressing its limitations in dataset diversity, adaptability to new AI implementations, or incorporation of multimodal imaging will all be important to maximize its utility. Future work should address the refinement of learning strategies that adapt the model to new data, increased representation in a variety of factors within the dataset, and the development of clinically relevant outcome metrics to establish ours as a tool in clinical, general dental applications.

\section*{Method} 

\subsection*{Data Collection and annotation} 
We constructed the largest paired CBCT-IOS dataset to date in collaboration with Delun Dental Hospital. CBCT images were acquired using a LargeV Smart3DX scanner (tube voltage: 100 kVp, tube current: 7 mA) with a slice thickness of 0.25 mm and an in-plane pixel spacing of 0.25 mm. IOS data were collected using a Shining3D Aoralscanner 3, which employs a non-contact, structured-light scanning principle at 20 frames per second and covers a scanning depth of -2 to 20 mm from the tip’s exit surface. Its illumination system incorporates green (520 nm), blue (448 nm), and white LED (400–780 nm) light sources. The dataset was compiled from 2022 to 2024, covering patients aged 15 to 86 who presented for dental treatment. The distribution of the dataset is illustrated in Figure~\ref{fig:overview}. Since our model employs a self-supervised contrastive learning framework, no manual annotations or labels were required.

\subsection*{Data Preprocessing} 
We performed data preprocessing for both CBCT and IOS data to standardize formats and ensure consistent anatomical orientation. For CBCT images, trilinear interpolation was used to resample each scan to 0.3 $\times$ 0.3 $\times$ 0.3\,$\text{mm}^3$ voxel spacing, followed by alignment to the RAS (Right, Anterior, Superior) orientation. We then clipped intensity values to [0, 2500] Hounsfield units and min-max normalized each volume to the [0, 1] range. Subsequently, a lightweight segmentation model was utilized to crop tooth regions, isolating the jaw from extraneous structures and reducing input data dimensions.
For IOS data, each jaw was collected in an occluded arrangement whenever possible. For scans that were not occluded at the time of collection, we employed a lightweight registration model to merge the upper and lower arches, standardizing occlusal contacts across all samples. We then applied farthest point sampling (FPS) to retain 24,000 points per jaw, as FPS preserves the global structure of the point cloud.
Finally, patient-level data splitting was carried out to prevent overlap among training, validation, and test sets, and all scans were anonymized in accordance with institutional guidelines.

\subsection*{CBCT-IOS Multimodal Contrastive Learning} 
Accurate representation and modeling of a patient's entire oral structure is paramount in dental treatment planning, particularly for digital dentistry involving dental prosthesis design and robotic interventions. Precise delineation of individual teeth from surrounding teeth and gingiva is crucial for creating an accurate digital representation and facilitating effective treatment planning. However, the limited availability of labeled data in CBCT and IOS necessitates label-efficient learning to leverage unlabeled data effectively. 

Self-supervised pretraining has emerged as an effective method for enhancing generalizability and performance in computational dental imaging by integrating CBCT and IOS. Prior research largely relied on unimodal approaches, limiting the potential to exploit complementary information from diverse modalities.

In this study, we propose a multimodal contrastive pretraining framework employing both cross-modal and intra-modal contrastive learning at the patch level. CBCT volumes undergo region-of-interest (ROI) extraction and are encoded via a Swin transformer. IOS data initially represented as meshes are converted into point cloud data (PCD) and processed through a Point Transformer (PT). Each modality generates a set of latent representations from their respective encoders. Intra-modal contrastive learning is applied within each modality to ensure consistency and robustness of modality-specific features. Simultaneously, cross-modal contrastive learning aligns latent representations from corresponding CBCT and IOS patches, enhancing cross-modal semantic consistency.

Our comprehensive multimodal contrastive pretraining paradigm effectively captures both modality-specific characteristics and cross-modal correlations. Extensive validation confirms our method's effectiveness and generalizability in clinical applications, demonstrating superior performance in both CBCT and IOS teeth segmentation.

Consider a multimodal dataset comprising $N$ paired CBCT-IOS scans, denoted as $\mathcal{D} = \{(C_1, P_1), (C_2, P_2), \ldots, (C_N, P_N)\}$, where $C_m$ represents the $m$-th CBCT image and $P_m$ represents the corresponding IOS point cloud. Our goal is to develop a multimodal dental vision model utilizing modality-specific encoders. Specifically, for a given CBCT-IOS pair $(C_m, P_m)$, a Swin Transformer-based image encoder $f(\cdot)$ processes the CBCT images, while a Point Transformer-based encoder $g(\cdot)$ processes the IOS point clouds, resulting in latent representations:

\begin{equation}
\mathbf{c}_{m} = f(C_m)
\end{equation}

\begin{equation}
\mathbf{p}_m = g(P_m)
\end{equation}

Here, $\mathbf{c}_m=\{\mathbf{c}_{m,1}, \mathbf{c}_{m,2},...,\mathbf{c}_{m,r}\}$ represents the encoded CBCT patch representations, where $\mathbf{c}_m\in\mathbb{R}^{r \times h}$. Likewise, $\mathbf{p}_m=\{\mathbf{p}_{m,1},\mathbf{p}_{m,2},...,\mathbf{p}_{m,s}\}$ represents the encoded IOS patch representations, where $\mathbf{p}_m \in \mathbb{R}^{s \times h}$. Here, $r$ and $s$ denote the number of patches in CBCT and IOS data respectively, $h$ is the hidden dimension. Patch-level contrastive learning is subsequently applied to these representations to ensure robust intra-modal and inter-modal semantic alignment.

\subsubsection*{Dual Patch-Level Alignment Strategy}

We propose a dual patch-level alignment framework that addresses two essential aspects:
(i) \textbf{Cross-Modal Alignment} (between CBCT and IOS), and
(ii) \textbf{Intra-Modal Alignment} (within each modality, CBCT or IOS), both operating at the patch level. This method effectively captures localized anatomical details, mitigating inaccuracies associated with exclusively global alignment methods.

\subsubsection*{Cross-Modal Patch Alignment}
Let $\mathcal{B}$ represent a mini-batch of paired CBCT and IOS patches. Utilizing a sigmoid-based contrastive loss, we enforce alignment between CBCT patches $\mathbf{c}_i$ and IOS patches $\mathbf{p}_j$:
\begin{equation}
    \label{eq:global}
    \mathcal{L}_{\text{Cross}}
    = -\frac{1}{|\mathcal{B}|} \sum_{i=1}^{|\mathcal{B}|} \sum_{j=1}^{|\mathcal{B}|}
      \log\!\Biggl(\frac{1}{1 + \exp\bigl( z^{ij}(-t_g\, \mathbf{c}_i \cdot \mathbf{p}_j + b_g) \bigr)}\Biggr),
\end{equation}
where $z^{ij} = 1$ for matched CBCT and IOS patch pairs, and $-1$ otherwise. Parameters $t_g$ and $b_g$ are the logit scale and bias, respectively. The embeddings $\mathbf{c}_i$ and $\mathbf{p}_j$ are normalized latent representations derived from the Swin Transformer (CBCT) and Point Transformer (IOS), facilitating consistent representation magnitudes.

\subsubsection*{Intra-Modal Patch Alignment}
In addition to cross-modal alignment, intra-modal alignment is performed within each modality to capture subtle structural variations. Let $\mathbf{p}_m$ represent the set of encoded patches from modality-specific encoders:
\begin{equation}
    \mathbf{p}_m = f(P_m),
\end{equation}
where $P_m$ is the raw patch input and $f(\cdot)$ the modality-specific encoder. To quantify the similarity between intra-modal patches, we form a similarity matrix $\mathbf{A}$:
\begin{equation}
    (\mathbf{a})_{ij} 
    = \frac{(\mathbf{p}_{m,i})^\top \mathbf{p}_{m,j}}{\|\mathbf{p}_{m,i}\|\;\|\mathbf{p}_{m,j}\|},
\end{equation}
\begin{equation}
    (\mathbf{A})_{ij} 
    = \log\!\biggl(\frac{1}{1 + \exp\bigl(z^{ij}(-t_l\,(\mathbf{a})_{ij} + b_l)\bigr)}\biggr),
\end{equation}
where $z^{ij} = 1$ if patches depict similar anatomical structures, otherwise $-1$; parameters $t_l$ and $b_l$ are the intra-modal logit scale and bias, respectively. The intra-modal alignment loss is then computed as:
\begin{equation}
    \label{eq:local}
    \mathcal{L}_{\text{Intra}}
    = -\frac{1}{|\mathcal{B}|} \sum_{m=1}^{|\mathcal{B}|}\sum_{n=1}^{|\mathcal{B}|} \frac{1}{w} \sum_{j=1}^{w} \max_{i} \bigl((\mathbf{A})_{ij}\bigr).
\end{equation}

\subsubsection*{Overall Objective}
Combining cross-modal and intra-modal alignment objectives, the final training loss is formulated as:
\begin{equation}
    \label{eq:total-loss}
    \mathcal{L}_{\text{Total}}
    = \mathcal{L}_{\text{Intra}}
    + \alpha \,\mathcal{L}_{\text{Cross}},
\end{equation}
where $\alpha$ balances the contributions from cross-modal and intra-modal alignments. This dual alignment framework ensures robust, anatomically-aware representations for both CBCT and IOS modalities, enhancing accuracy and generalization in downstream dental applications such as tooth segmentation and prosthetic planning.

\subsection*{CBCT and IOS Segmentation Model Fine-tuning} 
After pretraining our \ours{}, we fine-tuned it separately for each imaging modality to maximize segmentation performance. Modality-specific fine-tuning is necessary because CBCT and IOS data have fundamentally different characteristics. CBCT images are volumetric intensity data that often suffer from noise and low contrast artifacts, whereas IOS captures 3D surface geometry as unstructured point clouds. A pretrained model cannot optimally handle segmentations without adaptation. The pretrained backbone provides a strong initialization, but targeted fine-tuning on each modality’s labeled dataset is crucial to learn fine-grained tooth segmentation details.

\subsubsection*{Ground Truth}
The ground truth labels used for fine-tuning and evaluation were expert annotations provided with each public dataset. Labels were standardized to conform to the FDI numbering system when necessary. Metal artifacts were noted in some CBCT scans from the ToothFairy2 \cite{2024TMI} and Cui \cite{Cui2022FullyAutomaticAI} dataset.

\subsubsection*{CBCT Segmentation Fine-tuning}
For CBCT tooth segmentation, we adopted the SwinUNETR \cite{Hatamizadeh2022SwinUNETR} architecture as the segmentation model. SwinUNETR is a 3D U-Net variant that uses a Swin Transformer-based encoder, chosen over conventional CNNs due to its ability to capture long-range dependencies in volumetric data. Traditional convolutional networks have limited receptive fields, which can be suboptimal for structures like teeth that span varied scales and may be adjacent or overlapping in 3D space. In contrast, SwinUNETR’s encoder partitions the volume into non-overlapping 3D patches and processes them with hierarchical self-attention using shifted windows to model global context. This transformer encoder produces multi-scale feature maps, which are fed into a convolutional decoder via multiple skip connections, forming a U-Net style architecture. The hybrid design preserves fine local details through skip-connected high-resolution feature maps while also integrating global contextual information from the transformer encoder. 

Fine-tuning on CBCT was performed in a supervised manner with a dataset containing CBCT scans with fully labeled manual tooth annotations. Every tooth from the volume received its own respective label according to the FDI dental numbering system, which provides dental professionals with a coded representation of each tooth in one of four quadrants. For example, teeth located in the upper right quadrant are coded from 11-18, the upper left quadrant from 21-28, the lower left quadrant from 31-38, and the lower right quadrant from 41-48. The model was trained to classify types of tooth or teeth through multi-class segmentation to classify up to 33 class labels for teeth, metal artifacts, and background. The weights of the SwinUNETR encoder were initialized with the pretrained \ours{} parameters and randomly initialized the decoder weights. The loss function used for CBCT fine-tuning was a composite of Dice loss and cross-entropy loss.

\subsubsection*{IOS Segmentation Fine-tuning}
For the IOS tooth segmentation, we leveraged the Point Transformer network for point cloud segmentation. We chose a point-based transformer architecture as opposed to a voxel-based CNN since the IOS data is a point cloud - an irregular arrangement of points in 3D space that would have been cumbersome and would have lost geometric detail if converted to a voxel grid. The Point Transformer architecture uses self-attention layers to operate on point sets, clearly making it permutation invariant to the order of input points. This invariance is necessary by the nature of point clouds, and the order of points has no significance. The self-attention mechanism also enables the network to learn local neighborhood features in addition to long-range interactions between points, an important property for segmenting the detailed geometry of tooth surfaces and connections between teeth. In our application, we provide the Point Transformer with a set of points, where each point is defined by its coordinates in 3D and the normalization as the input. The Point Transformer then processes the points through blocks of vector self-attention and positional embedding, allowing it to aggregate information from nearby points and information globally across the shape. Through this process, the Point Transformer will output rich feature representations for each point that simultaneously incorporate both the local surface curvature and the overall arch structure.  

The IOS fine-tuning took the same supervised learning approach as the CBCT case. Our training dataset consisted of a set of IOS scans where each point in the point cloud was assigned a class corresponding to its tooth or the gingiva. We again used the FDI numbering scheme for tooth labels in the point cloud. The model would predict these per-point labels and detail the segmentation of the point cloud into distinct clusters for each individual tooth. The Point Transformer was fine-tuned on this task, initialized from the \ours{} pretrained weights. The loss function for IOS segmentation consisted of a point-wise cross-entropy loss.

\subsection*{Model Training and Implementation Details} 
All models were implemented using PyTorch \cite{paszke2019pytorch} and trained on 4 NVIDIA RTX 3090 or A800 GPUs . For the multimodal pretraining stage, augmentation strategies were applied at the patch level to introduce greater variability and improve model robustness. The \ours{} model was optimized with the AdamW optimizer \cite{Loshchilov2019Adamw} with an initial learning rate of $1 \times 10^{-4}$, a weight decay of $1 \times 10^{-5}$, and a cosine learning rate schedule with warmup \cite{Loshchilov2017SGDR}. The pretraining phase was conducted for 50 epochs, with the batch size determined by available GPU memory capacity to ensure stable training.

For modality-specific fine-tuning, the CBCT segmentation model (SwinUNETR \cite{Hatamizadeh2022SwinUNETR}) and the IOS segmentation model (PointTransformer \cite{zhao2021pointtransformer}) were optimized individually using the AdamW optimizer \cite{Loshchilov2019Adamw}. Both models maintained an initial learning rate of $3.0 \times 10^{-4}$ and a weight decay of $1 \times 10^{-5}$, following a cosine learning rate schedule with warmup \cite{Loshchilov2017SGDR}. Each model was trained for 150 epochs with a batch size of 1.

In the fine-tuning stage, CBCT images were randomly cropped to a consistent size of $256 \times 256 \times 128$. To maintain anatomical integrity and prevent label mismatches, data augmentations such as random flips were excluded. Instead, augmentations included random intensity shifts, random cropping, random gaussian noise, and random rotations within the range of $(-0.5, 0.5)$ radians. Similarly, mesh data augmentation for IOS involved random shifts, random rotations within the range of $(-0.5, 0.5)$, and random scaling.

\subsection*{Statistical Analysis}
We report a 95\% confidence interval (CI) for statistical analysis. All measurements are implemented with Python 3.9. Statistical significance was assessed at the 0.01 level unless otherwise specified, using two-sided paired t-tests where applicable.

\section*{Data Availability}
This study involved both private and public datasets for system development. The private data collected and processed in this study are supervised by the corresponding institutions. The data are available under restricted access, which can be obtained by emailing the corresponding author for academic use requests. The requirements will be evaluated in relation to institutional policies. Data can only be shared for non-commercial academic use with a formal material transfer agreement. All requests will be promptly reviewed within 15 working days. The public datasets comprises ToothFairy2 \cite{2024TMI}, Cui \cite{Cui2022FullyAutomaticAI}, Teeth3DS \cite{BenHamadou2022}, TADPM \cite{Lei2024AutomaticTADPM}, 3D-IOSSeg \cite{Li20243D-IOSSeg}. The public dataset was used for training and evaluating the downstream task. The details of the public dataset are shown in Extended Data Table.~\ref{table:datasets}.

\section*{Code Availability}
Codes will be available upon paper acceptance.

\section*{Author contribution}
M.S., J.B., Z.Q., Y.L., and H.C. conceptualized and designed the study. M.S., J.B., and K.L. collected and curated the \ourdataset{} dataset, performing data preprocessing and quality control. M.S. developed and implemented the pretraining algorithm and drafted the manuscript. J.P. and Y.L. provided clinical guidance and domain-specific expertise throughout the study. H.C. supervised the project and contributed to revising the manuscript.

\section*{Illustrations}
Plots were created using Python matplotlib library. Portions of the illustration in Figure~\ref{fig:overview} were generated using OpenAI's GPT-4o.

\section*{Declarations}
The authors have no conflicts of interest to declare. Kai Xin Li is affiliated with Delun Dental Hospital, Guangzhou, China. This affiliation did not influence the study design, data interpretation, or decision to publish.

\section*{Ethics approval}
This study has been reviewed and approved by the Human and Artefacts Research Ethics Committee (HAREC). The protocol number is HREP-2024-0257.

\section*{Acknowledgements}
This work was supported by the Hong Kong Innovation and Technology Commission (Project No. GHP/006/22GD and ITCPD/17-9), and the Research Grants Council of the Hong Kong Special Administrative Region, China (Project No. T45-401/22-N). We thank Chen Xiao Juan (a technician at Delun Dental Hospital) and Dr. Jiang Tie Long (a technician at Shenzhen Eno-Dental Clinic) for their valuable assistance in providing the external datasets used in this study.
\newpage
\bibliography{sn-bibliography}

\begin{thebibliography}{10}
\expandafter\ifx\csname url\endcsname\relax
  \def\url#1{\burl{#1}}\fi
\expandafter\ifx\csname urlprefix\endcsname\relax\def\urlprefix{URL }\fi
\providecommand{\bibinfo}[2]{#2}
\providecommand{\eprint}[2][]{\url{#2}}
\providecommand{\doi}[1]{\url{https://doi.org/#1}}
\bibcommenthead

\bibitem{WHOOralHealth2022}
\bibinfo{author}{{World Health Organization}}.
\newblock \bibinfo{title}{Global oral health status report: Towards universal health coverage for oral health by 2030} (\bibinfo{year}{2022}).

\bibitem{william2006clinicalCBCT}
\bibinfo{author}{Scarfe, W.~C.}, \bibinfo{author}{Farman, A.~G.} \& \bibinfo{author}{Suković, M.}
\newblock \bibinfo{title}{Clinical applications of cone-beam computed tomography in dental practice}.
\newblock \emph{\bibinfo{journal}{Journal of the Canadian Dental Association}} \textbf{\bibinfo{volume}{72}}, \bibinfo{pages}{75--80} (\bibinfo{year}{2006}).

\bibitem{Mangano2017Intraoral}
\bibinfo{author}{Mangano, F.}, \bibinfo{author}{Gandolfi, A.}, \bibinfo{author}{Luongo, G.} \& \bibinfo{author}{Logozzo, S.}
\newblock \bibinfo{title}{Intraoral scanners in dentistry: a review of the current literature}.
\newblock \emph{\bibinfo{journal}{BMC Oral Health}} \textbf{\bibinfo{volume}{17}}, \bibinfo{pages}{149} (\bibinfo{year}{2017}).

\bibitem{Baldini2025}
\bibinfo{author}{Baldini, B.}, \bibinfo{author}{Papasratorn, D.}, \bibinfo{author}{Fagundes, F.~B.}, \bibinfo{author}{Fontenele, R.~C.} \& \bibinfo{author}{Jacobs, R.}
\newblock \bibinfo{title}{Validation of a novel tool for automated tooth modelling by fusion of cbct-derived roots with the respective ios-derived crowns}.
\newblock \emph{\bibinfo{journal}{Journal of Dentistry}} \textbf{\bibinfo{volume}{153}}, \bibinfo{pages}{105546} (\bibinfo{year}{2025}).

\bibitem{Albano2024Caries}
\bibinfo{author}{Albano, D.} \emph{et~al.}
\newblock \bibinfo{title}{Artificial intelligence for radiographic imaging detection of caries lesions: a systematic review}.
\newblock \emph{\bibinfo{journal}{BMC Oral Health}} \textbf{\bibinfo{volume}{24}}, \bibinfo{pages}{274} (\bibinfo{year}{2024}).

\bibitem{Adnan2024Caries}
\bibinfo{author}{Adnan, N.} \emph{et~al.}
\newblock \bibinfo{title}{Developing an ai-based application for caries index detection on intraoral photographs}.
\newblock \emph{\bibinfo{journal}{Scientific Reports}} \textbf{\bibinfo{volume}{14}}, \bibinfo{pages}{26752} (\bibinfo{year}{2024}).

\bibitem{Negi2024Caries}
\bibinfo{author}{Negi, S.} \emph{et~al.}
\newblock \bibinfo{title}{Artificial intelligence in dental caries diagnosis and detection: An umbrella review}.
\newblock \emph{\bibinfo{journal}{Clinical and Experimental Dental Research}} \textbf{\bibinfo{volume}{10}}, \bibinfo{pages}{e70004} (\bibinfo{year}{2024}).

\bibitem{Liu2023Orthodontics}
\bibinfo{author}{Liu, J.}, \bibinfo{author}{Zhang, C.} \& \bibinfo{author}{Shan, Z.}
\newblock \bibinfo{title}{Application of artificial intelligence in orthodontics: Current state and future perspectives}.
\newblock \emph{\bibinfo{journal}{Healthcare}} \textbf{\bibinfo{volume}{11}}, \bibinfo{pages}{2760} (\bibinfo{year}{2023}).

\bibitem{Deng2024TAPoseNet}
\bibinfo{author}{Deng, Q.}, \bibinfo{author}{Yang, X.}, \bibinfo{author}{Huang, M.}, \bibinfo{author}{Jiang, L.} \& \bibinfo{author}{Zhang, D.} \emph{\bibinfo{title}{Taposenet: Teeth alignment based on pose estimation via multi-scale graph convolutional network}}.
\newblock \emph{\bibinfo{booktitle}{Proceedings of Medical Image Computing and Computer Assisted Intervention -- MICCAI 2024}}, Vol. \bibinfo{volume}{15012}, \bibinfo{pages}{314--323} (\bibinfo{publisher}{Springer Nature Switzerland}, \bibinfo{year}{2024}).

\bibitem{Lei2024AutomaticTADPM}
\bibinfo{author}{Lei, C.} \emph{et~al.}
\newblock \bibinfo{title}{Automatic tooth arrangement with joint features of point and mesh representations via diffusion probabilistic models}.
\newblock \emph{\bibinfo{journal}{Computer Aided Geometric Design}} \textbf{\bibinfo{volume}{111}}, \bibinfo{pages}{102293} (\bibinfo{year}{2024}).

\bibitem{Kong2024Prosthesis}
\bibinfo{author}{Kong, H.-J.} \& \bibinfo{author}{Kim, Y.-L.}
\newblock \bibinfo{title}{Application of artificial intelligence in dental crown prosthesis: A scoping review}.
\newblock \emph{\bibinfo{journal}{BMC Oral Health}} \textbf{\bibinfo{volume}{24}}, \bibinfo{pages}{937} (\bibinfo{year}{2024}).

\bibitem{Hosseinimanesh2023AICrown}
\bibinfo{author}{Hosseinimanesh, G.}, \bibinfo{author}{Ghadiri, F.}, \bibinfo{author}{Guibault, F.}, \bibinfo{author}{Cheriet, F.} \& \bibinfo{author}{Keren, J.} \emph{\bibinfo{title}{From mesh completion to ai designed crown}}.
\newblock \emph{\bibinfo{booktitle}{Proceedings of Medical Image Computing and Computer-Assisted Intervention -- MICCAI 2023}}, Vol. \bibinfo{volume}{12905} of \emph{\bibinfo{series}{Lecture Notes in Computer Science}}, \bibinfo{pages}{543--552} (\bibinfo{publisher}{Springer}, \bibinfo{year}{2023}).

\bibitem{Chau2024SingleMolar}
\bibinfo{author}{Chau, R. C.~W.}, \bibinfo{author}{Hsung, R. T.-C.}, \bibinfo{author}{Koohi-Moghadam, M.} \& \bibinfo{author}{Lam, W. Y.~H.}
\newblock \bibinfo{title}{Accuracy of artificial intelligence-designed single-molar dental prostheses: A feasibility study}.
\newblock \emph{\bibinfo{journal}{The Journal of Prosthetic Dentistry}} \textbf{\bibinfo{volume}{131}}, \bibinfo{pages}{1111--1117} (\bibinfo{year}{2024}).

\bibitem{Chen2024Segmentation}
\bibinfo{author}{Chen, X.}, \bibinfo{author}{Ma, N.}, \bibinfo{author}{Xu, T.} \& \bibinfo{author}{Xu, C.}
\newblock \bibinfo{title}{Deep learning-based tooth segmentation methods in medical imaging: A review}.
\newblock \emph{\bibinfo{journal}{Proceedings of the Institution of Mechanical Engineers, Part H: Journal of Engineering in Medicine}} \textbf{\bibinfo{volume}{238}}, \bibinfo{pages}{3--20} (\bibinfo{year}{2024}).

\bibitem{lecun1998cnn}
\bibinfo{author}{LeCun, Y.}, \bibinfo{author}{Bottou, L.}, \bibinfo{author}{Bengio, Y.} \& \bibinfo{author}{Haffner, P.}
\newblock \bibinfo{title}{Gradient-based learning applied to document recognition}.
\newblock \emph{\bibinfo{journal}{Proceedings of the IEEE}} \textbf{\bibinfo{volume}{86}}, \bibinfo{pages}{2278--2324} (\bibinfo{year}{1998}).

\bibitem{ronneberger2015unet}
\bibinfo{author}{Ronneberger, O.}, \bibinfo{author}{Fischer, P.} \& \bibinfo{author}{Brox, T.} \emph{\bibinfo{title}{U-net: Convolutional networks for biomedical image segmentation}}.
\newblock \emph{\bibinfo{booktitle}{Medical Image Computing and Computer-Assisted Intervention -- MICCAI 2015}}, \bibinfo{pages}{234--241} (\bibinfo{organization}{Springer International Publishing}, \bibinfo{year}{2015}).

\bibitem{Cui2022FullyAutomaticAI}
\bibinfo{author}{Cui, Z.} \emph{et~al.}
\newblock \bibinfo{title}{A fully automatic ai system for tooth and alveolar bone segmentation from cone-beam ct images}.
\newblock \emph{\bibinfo{journal}{Nature Communications}} \textbf{\bibinfo{volume}{13}}, \bibinfo{pages}{2096} (\bibinfo{year}{2022}).

\bibitem{vaswani2017attention}
\bibinfo{author}{Vaswani, A.} \emph{et~al.} \emph{\bibinfo{title}{Attention is all you need}}.
\newblock \emph{\bibinfo{booktitle}{Advances in Neural Information Processing Systems}}, \bibinfo{pages}{5998--6008} (\bibinfo{year}{2017}).

\bibitem{Gillot2022AMASS}
\bibinfo{author}{Gillot, M.} \emph{et~al.}
\newblock \bibinfo{title}{Automatic multi-anatomical skull structure segmentation of cone-beam computed tomography scans using 3d unetr}.
\newblock \emph{\bibinfo{journal}{PLOS ONE}} \textbf{\bibinfo{volume}{17}}, \bibinfo{pages}{e0275033} (\bibinfo{year}{2022}).

\bibitem{Gu2023mamba}
\bibinfo{author}{Gu, A.} \& \bibinfo{author}{Dao, T.} \emph{\bibinfo{title}{Mamba: Linear-time sequence modeling with selective state spaces}}.
\newblock \emph{\bibinfo{booktitle}{Proceedings of the Conference on Language Modeling (COLM)}} (\bibinfo{year}{2025}).

\bibitem{hao2024tmamba}
\bibinfo{author}{Hao, J.} \emph{et~al.}
\newblock \bibinfo{title}{T-mamba: A unified framework with long-range dependency in dual-domain for 2d \& 3d tooth segmentation}.
\newblock \emph{\bibinfo{journal}{arXiv preprint arXiv:2404.01065}}  (\bibinfo{year}{2024}).

\bibitem{scarselli2009gnn}
\bibinfo{author}{Scarselli, F.}, \bibinfo{author}{Gori, M.}, \bibinfo{author}{Tsoi, A.~C.}, \bibinfo{author}{Hagenbuchner, M.} \& \bibinfo{author}{Monfardini, G.}
\newblock \bibinfo{title}{The graph neural network model}.
\newblock \emph{\bibinfo{journal}{IEEE Transactions on Neural Networks}} \textbf{\bibinfo{volume}{20}}, \bibinfo{pages}{61--80} (\bibinfo{year}{2009}).

\bibitem{BenHamadou2022}
\bibinfo{author}{Ben-Hamadou, A.}, \bibinfo{author}{Smaoui, O.}, \bibinfo{author}{Chaabouni-Chouayakh, H.} \& \bibinfo{author}{Mahjoub, M.~A.}
\newblock \bibinfo{title}{Teeth3ds: A benchmark for teeth segmentation and labeling from intra-oral 3d scans}.
\newblock \emph{\bibinfo{journal}{arXiv preprint arXiv:2210.06094}}  (\bibinfo{year}{2022}).

\bibitem{ISO3950_2016}
\bibinfo{author}{{International Organization for Standardization}}.
\newblock \bibinfo{title}{{Dentistry—Designation system for teeth and areas of the oral cavity}}.
\newblock \bibinfo{howpublished}{ISO 3950:2016} (\bibinfo{year}{2016}).

\bibitem{Schwendicke2020Artificial}
\bibinfo{author}{Schwendicke, F.}, \bibinfo{author}{Samek, W.} \& \bibinfo{author}{Krois, J.}
\newblock \bibinfo{title}{Artificial intelligence in dentistry: Chances and challenges}.
\newblock \emph{\bibinfo{journal}{Journal of Dental Research}} \textbf{\bibinfo{volume}{99}}, \bibinfo{pages}{769--774} (\bibinfo{year}{2020}).

\bibitem{chen2020Contrastive}
\bibinfo{author}{Chen, T.}, \bibinfo{author}{Kornblith, S.}, \bibinfo{author}{Norouzi, M.} \& \bibinfo{author}{Hinton, G.} \emph{\bibinfo{title}{A simple framework for contrastive learning of visual representations}}.
\newblock \emph{\bibinfo{booktitle}{Proceedings of the 37th International Conference on Machine Learning (ICML)}}, \bibinfo{pages}{1597--1607} (\bibinfo{year}{2020}).

\bibitem{radford2021Clip}
\bibinfo{author}{Radford, A.} \emph{et~al.} \emph{\bibinfo{title}{Learning transferable visual models from natural language supervision}}.
\newblock \emph{\bibinfo{booktitle}{Proceedings of the 38th International Conference on Machine Learning (ICML)}} (\bibinfo{year}{2021}).

\bibitem{Tang2022SelfSupervised}
\bibinfo{author}{Tang, Y.} \emph{et~al.} \emph{\bibinfo{title}{Self-supervised pre-training of swin transformers for 3d medical image analysis}}.
\newblock \emph{\bibinfo{booktitle}{Proceedings of the IEEE/CVF Conference on Computer Vision and Pattern Recognition}}, \bibinfo{pages}{20730--20740} (\bibinfo{year}{2022}).

\bibitem{Isensee2021nnUNet}
\bibinfo{author}{Isensee, F.}, \bibinfo{author}{Jaeger, P.~F.}, \bibinfo{author}{Kohl, S. A.~A.}, \bibinfo{author}{Petersen, J.} \& \bibinfo{author}{Maier-Hein, K.~H.}
\newblock \bibinfo{title}{nnu-net: a self-configuring method for deep learning-based biomedical image segmentation}.
\newblock \emph{\bibinfo{journal}{Nature Methods}} \textbf{\bibinfo{volume}{18}}, \bibinfo{pages}{203--211} (\bibinfo{year}{2021}).

\bibitem{qi2017pointnet}
\bibinfo{author}{Qi, C.~R.}, \bibinfo{author}{Su, H.}, \bibinfo{author}{Mo, K.} \& \bibinfo{author}{Guibas, L.~J.} \emph{\bibinfo{title}{Pointnet: Deep learning on point sets for 3d classification and segmentation}}.
\newblock \emph{\bibinfo{booktitle}{Proceedings of the IEEE Conference on Computer Vision and Pattern Recognition (CVPR)}}, \bibinfo{pages}{652--660} (\bibinfo{year}{2017}).

\bibitem{wang2019dgcnn}
\bibinfo{author}{Wang, Y.} \emph{et~al.}
\newblock \bibinfo{title}{Dynamic graph cnn for learning on point clouds}.
\newblock \emph{\bibinfo{journal}{ACM Transactions on Graphics (TOG)}}  (\bibinfo{year}{2019}).

\bibitem{lee2023tsegformer}
\bibinfo{author}{Xiong, H.} \emph{et~al.} \emph{\bibinfo{title}{Tsegformer: 3d tooth segmentation in intraoral scans with geometry guided transformer}}.
\newblock \emph{\bibinfo{booktitle}{Proceedings of Medical Image Computing and Computer Assisted Intervention -- MICCAI 2023}} (\bibinfo{year}{2023}).

\bibitem{2024TMI}
\bibinfo{author}{Bolelli, F.} \emph{et~al.}
\newblock \bibinfo{title}{{Segmenting the Inferior Alveolar Canal in CBCTs Volumes: the ToothFairy Challenge}}.
\newblock \emph{\bibinfo{journal}{IEEE Transactions on Medical Imaging}} \bibinfo{pages}{1--17} (\bibinfo{year}{2024}).

\bibitem{Hatamizadeh2022SwinUNETR}
\bibinfo{author}{Hatamizadeh, A.} \emph{et~al.} \emph{\bibinfo{title}{Swin unetr: Swin transformers for semantic segmentation of brain tumors in mri images}}.
\newblock \emph{\bibinfo{booktitle}{Brainlesion: Glioma, Multiple Sclerosis, Stroke and Traumatic Brain Injuries}}, Vol. \bibinfo{volume}{12962} of \emph{\bibinfo{series}{Lecture Notes in Computer Science}}, \bibinfo{pages}{272--284} (\bibinfo{year}{2022}).

\bibitem{paszke2019pytorch}
\bibinfo{author}{Paszke, A.} \emph{et~al.} \emph{\bibinfo{title}{Pytorch: An imperative style, high-performance deep learning library}}.
\newblock \emph{\bibinfo{booktitle}{Advances in Neural Information Processing Systems 32}}, \bibinfo{pages}{8024--8035} (\bibinfo{year}{2019}).

\bibitem{Loshchilov2019Adamw}
\bibinfo{author}{Loshchilov, I.} \& \bibinfo{author}{Hutter, F.} \emph{\bibinfo{title}{Decoupled weight decay regularization}}.
\newblock \emph{\bibinfo{booktitle}{International Conference on Learning Representations}} (\bibinfo{year}{2019}).

\bibitem{Loshchilov2017SGDR}
\bibinfo{author}{Loshchilov, I.} \& \bibinfo{author}{Hutter, F.} \emph{\bibinfo{title}{{SGDR}: Stochastic gradient descent with warm restarts}}.
\newblock \emph{\bibinfo{booktitle}{International Conference on Learning Representations (ICLR)}} (\bibinfo{year}{2017}).

\bibitem{zhao2021pointtransformer}
\bibinfo{author}{Zhao, H.}, \bibinfo{author}{Jiang, L.}, \bibinfo{author}{Jia, J.}, \bibinfo{author}{Torr, P. H.~S.} \& \bibinfo{author}{Koltun, V.} \emph{\bibinfo{title}{Point transformer}}.
\newblock \emph{\bibinfo{booktitle}{Proceedings of the IEEE/CVF International Conference on Computer Vision (ICCV)}}, \bibinfo{pages}{16259--16268} (\bibinfo{year}{2021}).

\bibitem{Li20243D-IOSSeg}
\bibinfo{author}{Li, J.} \emph{et~al.}
\newblock \bibinfo{title}{A fine-grained orthodontics segmentation model for 3d intraoral scan data}.
\newblock \emph{\bibinfo{journal}{Computers in Biology and Medicine}} \textbf{\bibinfo{volume}{168}}, \bibinfo{pages}{107821} (\bibinfo{year}{2024}).

\bibitem{2022CVPR}
\bibinfo{author}{Cipriano, M.}, \bibinfo{author}{Allegretti, S.}, \bibinfo{author}{Bolelli, F.}, \bibinfo{author}{Pollastri, F.} \& \bibinfo{author}{Grana, C.} \emph{\bibinfo{title}{{Improving Segmentation of the Inferior Alveolar Nerve through Deep Label Propagation}}}.
\newblock \emph{\bibinfo{booktitle}{IEEE/CVF Conference on Computer Vision and Pattern Recognition (CVPR)}}, \bibinfo{pages}{21105--21114} (\bibinfo{publisher}{IEEE}, \bibinfo{year}{2022}).

\end{thebibliography}
\clearpage
\begin{appendices}

\section{Extended Data}\label{secA1}
\fontsize{8}{11}\selectfont
\begin{longtable}{|p{.15\textwidth}|p{.23\textwidth}|p{.52\textwidth}|}
\caption{Description of Used Dental CBCT and IOS Datasets.}
\label{table:datasets} \\
\toprule
Dataset & Website & Description \\
\midrule
\endfirsthead
\multicolumn{3}{c}{{\bfseries \tablename\ \thetable{} -- continued from previous page}} \\
\toprule
Dataset & Website & Description \\
\midrule
\endhead
\midrule
\multicolumn{3}{r}{{Continued on next page}} \\
\bottomrule
\endfoot
\bottomrule
\endlastfoot
\multicolumn{3}{|c|}{\textbf{CBCT-IOS Paired}} \\
\midrule
\ourdataset{} & Private & The dataset comprises 3,867 paired CBCT and IOS. No labels are available. \\ 
\midrule
\multicolumn{3}{|c|}{\textbf{CBCT Segmentation}} \\
\midrule
ToothFairy2 {\cite{2024TMI}} & \url{https://toothfairy2.grand-challenge.org/} & The dataset comprises 480 fully labeled CBCT. \\ 
\midrule
Cui {\cite{Cui2022FullyAutomaticAI}} & \url{https://github.com/ErdanC/Tooth-and-alveolar-bone-segmentation-from-CBCT} & The dataset comprises 148 fully labeled CBCT \\
\midrule
\multicolumn{3}{|c|}{\textbf{IOS Segmentation}} \\
\midrule
Teeth3DS {\cite{BenHamadou2022}} & \url{https://osf.io/xctdy/} & The dataset comprises 900 fully labeled IOS \\
\midrule
TADPM {\cite{Lei2024AutomaticTADPM}} & \url{https://github.com/lcshhh/TADPM/} & The dataset comprises 2,120 fully labeled IOS, corresponding to 1,060 cases with paired pre- and post-orthodontic treatment scans.\\
\midrule
3D-IOSSeg {\cite{Li20243D-IOSSeg}} & \url{https://github.com/MIVRC/Fast-TGCN-Pytorch-main} & The dataset comprises 90 fully labeled IOS \\
\end{longtable}
\clearpage
\begin{sidewaystable*}[h]
    \centering
    \caption{Comparison of CBCT tooth segmentation performance across different methods and datasets. Dice score (\%) is reported as mean (95\% confidence interval). Bold values indicate the best performance.}
    \label{table:cbct_results}
    \resizebox{\textwidth}{!}{
    \begin{tabular}{c|c|c|c|c|c|c|c|c|c|c}
        \multirow{2}{*}{}                                           & \multirow{2}{*}{Method} & \multirow{2}{*}{Total ↑}      & T1 ↑                          & T2 ↑                          & T3 ↑                          & T4 ↑                          & T5 ↑                            & T6 ↑                          & T7 ↑                            & T8 ↑                            \\
                                                                    &                        &                               & (Central Incisor)             & (Lateral   Incisor)           & (Cuspid)                      & (1st   Premolar)              & (2nd   Premolar)                & (1st   Molar)                 & (2nd   Molar)                   & (3rd   Molar)                   \\ \hline
        \multirow{4}{*}[1.5ex]{\rotatebox[origin=c]{90}{\parbox{3cm}{%
            \centering ToothFairy2 {\cite{2024TMI}} \\ (Internal)}}} & Ours                                          & \textbf{\begin{tabular}[c]{@{}c@{}}93.38 \\ (92.02, 94.73)\end{tabular}} & \textbf{\begin{tabular}[c]{@{}c@{}}96.21 \\ (95.34, 97.08)\end{tabular}} & \textbf{\begin{tabular}[c]{@{}c@{}}96.10 \\ (95.03, 97.17)\end{tabular}} & \textbf{\begin{tabular}[c]{@{}c@{}}94.37 \\ (92.01, 96.74)\end{tabular}} & \textbf{\begin{tabular}[c]{@{}c@{}}92.33 \\ (90.43, 94.23)\end{tabular}} & \begin{tabular}[c]{@{}c@{}}92.83 \\ (90.90, 94.76)\end{tabular}            & \textbf{\begin{tabular}[c]{@{}c@{}}92.14 \\ (89.93, 94.36)\end{tabular}} & \begin{tabular}[c]{@{}c@{}}91.56 \\ (89.31, 93.80)\end{tabular}            & \begin{tabular}[c]{@{}c@{}}92.92 \\ (90.28, 95.55)\end{tabular}            \\
                                                                    & SSL   {\cite{2022CVPR}}                       & \begin{tabular}[c]{@{}c@{}}90.69 \\ (88.91, 92.46)\end{tabular}        & \begin{tabular}[c]{@{}c@{}}94.08 \\ (92.36, 95.80)\end{tabular}        & \begin{tabular}[c]{@{}c@{}}94.81 \\ (93.13, 96.48)\end{tabular}        & \begin{tabular}[c]{@{}c@{}}93.59 \\ (90.96, 96.23)\end{tabular}        & \begin{tabular}[c]{@{}c@{}}89.75 \\ (86.86, 92.63)\end{tabular}        & \begin{tabular}[c]{@{}c@{}}92.24 \\ (90.01, 94.47)\end{tabular}          & \begin{tabular}[c]{@{}c@{}}85.57 \\ (81.10, 90.05)\end{tabular}        & \begin{tabular}[c]{@{}c@{}}86.77 \\ (83.21, 90.33)\end{tabular}          & \begin{tabular}[c]{@{}c@{}}88.37 \\ (84.80, 91.95)\end{tabular}          \\
                                                                    & nnUNet   {\cite{Isensee2021nnUNet}}           & \begin{tabular}[c]{@{}c@{}}82.09 \\ (78.07, 86.11)\end{tabular}        & \begin{tabular}[c]{@{}c@{}}89.32 \\ (85.11, 93.53)\end{tabular}        & \begin{tabular}[c]{@{}c@{}}90.10 \\ (86.09, 94.11)\end{tabular}        & \begin{tabular}[c]{@{}c@{}}87.07 \\ (82.44, 91.69)\end{tabular}        & \begin{tabular}[c]{@{}c@{}}81.91 \\ (76.63, 87.19)\end{tabular}        & \begin{tabular}[c]{@{}c@{}}81.69 \\ (76.08, 87.29)\end{tabular}          & \begin{tabular}[c]{@{}c@{}}77.93 \\ (72.30, 83.56)\end{tabular}        & \begin{tabular}[c]{@{}c@{}}69.20 \\ (62.42, 75.98)\end{tabular}          & \begin{tabular}[c]{@{}c@{}}67.08 \\ (59.86, 74.29)\end{tabular}          \\
                                                                    & Cui {\cite{Cui2022FullyAutomaticAI}}   & \begin{tabular}[c]{@{}c@{}}88.08 \\ (86.45, 89.72)\end{tabular}        & \begin{tabular}[c]{@{}c@{}}70.82 \\ (66.85, 74.79)\end{tabular}        & \begin{tabular}[c]{@{}c@{}}95.16 \\ (94.32, 96.00)\end{tabular}        & \begin{tabular}[c]{@{}c@{}}92.82 \\ (90.76, 94.89)\end{tabular}        & \begin{tabular}[c]{@{}c@{}}90.27 \\ (87.70, 92.84)\end{tabular}        & \textbf{\begin{tabular}[c]{@{}c@{}}93.76 \\ (92.13, 95.39)\end{tabular}} & \begin{tabular}[c]{@{}c@{}}91.64 \\ (88.69, 94.59)\end{tabular}        & \textbf{\begin{tabular}[c]{@{}c@{}}94.31 \\ (93.01, 95.62)\end{tabular}} & \textbf{\begin{tabular}[c]{@{}c@{}}92.95 \\ (89.63, 96.28)\end{tabular}} \\ \hline
        \multirow{4}{*}[1.5ex]{\rotatebox[origin=c]{90}{\parbox{3cm}{%
            \centering Cui {\cite{Cui2022FullyAutomaticAI}} \\ (External)}}}    & Ours                           & \textbf{\begin{tabular}[c]{@{}c@{}}88.92 \\ (87.97, 89.87)\end{tabular}} & \textbf{\begin{tabular}[c]{@{}c@{}}88.10 \\ (86.30, 89.90)\end{tabular}} & \textbf{\begin{tabular}[c]{@{}c@{}}89.25 \\ (88.14, 90.36)\end{tabular}} & \textbf{\begin{tabular}[c]{@{}c@{}}91.32 \\ (89.84, 92.79)\end{tabular}} & \textbf{\begin{tabular}[c]{@{}c@{}}89.89 \\ (88.77, 91.00)\end{tabular}} & \textbf{\begin{tabular}[c]{@{}c@{}}87.67 \\ (86.12, 89.22)\end{tabular}}   & \textbf{\begin{tabular}[c]{@{}c@{}}90.81 \\ (90.09, 91.54)\end{tabular}} & \textbf{\begin{tabular}[c]{@{}c@{}}90.39 \\ (89.39, 91.39)\end{tabular}}   & \textbf{\begin{tabular}[c]{@{}c@{}}88.85 \\ (87.27, 90.42)\end{tabular}}   \\
                                                                    & SSL   {\cite{2022CVPR}}                       & \begin{tabular}[c]{@{}c@{}}79.07 \\ (77.27, 80.87)\end{tabular}        & \begin{tabular}[c]{@{}c@{}}83.21 \\ (81.26, 85.16)\end{tabular}        & \begin{tabular}[c]{@{}c@{}}82.71 \\ (80.81, 84.61)\end{tabular}        & \begin{tabular}[c]{@{}c@{}}85.47 \\ (83.93, 87.01)\end{tabular}        & \begin{tabular}[c]{@{}c@{}}80.57 \\ (78.38, 82.77)\end{tabular}        & \begin{tabular}[c]{@{}c@{}}77.14 \\ (74.42, 79.86)\end{tabular}          & \begin{tabular}[c]{@{}c@{}}79.11 \\ (75.89, 82.33)\end{tabular}        & \begin{tabular}[c]{@{}c@{}}74.03 \\ (71.08, 76.98)\end{tabular}          & \begin{tabular}[c]{@{}c@{}}71.44 \\ (67.43, 75.44)\end{tabular}          \\
                                                                    & nnUNet   {\cite{Isensee2021nnUNet}}           & \begin{tabular}[c]{@{}c@{}}72.25 \\ (70.63, 73.87)\end{tabular}        & \begin{tabular}[c]{@{}c@{}}73.95 \\ (71.54 76.35)\end{tabular}        & \begin{tabular}[c]{@{}c@{}}70.41 \\ (68.08, 72.74)\end{tabular}        & \begin{tabular}[c]{@{}c@{}}73.27 \\ (71.22, 75.32)\end{tabular}        & \begin{tabular}[c]{@{}c@{}}72.41 \\ (70.29, 74.53)\end{tabular}        & \begin{tabular}[c]{@{}c@{}}68.41 \\ (65.86, 70.95)\end{tabular}          & \begin{tabular}[c]{@{}c@{}}73.65 \\ (71.51, 75.79)\end{tabular}        & \begin{tabular}[c]{@{}c@{}}72.51 \\ (69.85, 75.17)\end{tabular}          & \begin{tabular}[c]{@{}c@{}}77.22 \\ (74.54, 79.89)\end{tabular}          \\
                                                                    & Cui {\cite{Cui2022FullyAutomaticAI}}   & \begin{tabular}[c]{@{}c@{}}75.44 \\ (73.97, 76.90)\end{tabular}        & \begin{tabular}[c]{@{}c@{}}77.37 \\ (74.97, 79.77)\end{tabular}        & \begin{tabular}[c]{@{}c@{}}75.58 \\ (72.56, 78.69)\end{tabular}        & \begin{tabular}[c]{@{}c@{}}86.66 \\ (84.78, 88.54)\end{tabular}        & \begin{tabular}[c]{@{}c@{}}86.76 \\ (85.45, 88.07)\end{tabular}        & \begin{tabular}[c]{@{}c@{}}87.53 \\ (85.95, 89.12)\end{tabular}          & \begin{tabular}[c]{@{}c@{}}62.67 \\ (57.43, 67.91)\end{tabular}        & \begin{tabular}[c]{@{}c@{}}61.12 \\ (56.42, 65.81)\end{tabular}          & \begin{tabular}[c]{@{}c@{}}53.47 \\ (48.88, 58.07)\end{tabular}         
    \end{tabular}}
\end{sidewaystable*}
\clearpage
\begin{sidewaystable*}[h]
    \centering
    \caption{Comparison of IOS tooth segmentation performance across different methods and datasets. Dice score (\%) is reported as mean (95\% confidence interval). Bold values indicate the best performance.}
    \label{table:segmentation_results}
    \resizebox{\textwidth}{!}{%
    \begin{tabular}{c|c|c|c|c|c|c|c|c|c|c}
        \multirow{2}{*}{}                             & \multirow{2}{*}{Method} & \multirow{2}{*}{Total ↑}      & T1 ↑                            & T2 ↑                            & T3 ↑                            & T4 ↑                          & T5 ↑                                                            & T6 ↑                          & T7 ↑                          & T8 ↑                            \\
                                                      &                           &                               & (Central Incisor)               & (Lateral   Incisor)             & (Cuspid)                        & (1st   Premolar)              & (2nd   Premolar)                                                & (1st   Molar)                 & (2nd   Molar)                 & (3rd   Molar)                   \\ \hline
                \multirow{5}{*}[1.5ex]{\rotatebox[origin=c]{90}{\parbox{4cm}{%
            \centering Teeth3DS {\cite{BenHamadou2022}} (Internal)}}}      & Ours                      & \begin{tabular}[c]{@{}c@{}}\textbf{93.16} \\ \textbf{(92.58, 93.74)}\end{tabular} & \begin{tabular}[c]{@{}c@{}}\textbf{93.61} \\ \textbf{(92.79, 94.44)}\end{tabular}   & \begin{tabular}[c]{@{}c@{}}\textbf{93.42} \\ \textbf{(92.67, 94.17)}\end{tabular} & \begin{tabular}[c]{@{}c@{}}\textbf{92.71} \\ \textbf{(91.68, 93.73)}\end{tabular} & \begin{tabular}[c]{@{}c@{}}{92.70} \\ {(91.75, 93.65)}\end{tabular}            & \begin{tabular}[c]{@{}c@{}}\textbf{92.63} \\ \textbf{(91.72, 93.54)}\end{tabular} & \begin{tabular}[c]{@{}c@{}}\textbf{94.22} \\ \textbf{(93.44, 95.00)}\end{tabular} & \begin{tabular}[c]{@{}c@{}}{91.20} \\ {(89.85, 92.55)}\end{tabular} & \begin{tabular}[c]{@{}c@{}}{84.65} \\ {(82.82, 86.49)}\end{tabular} \\
                                                              & TGNet   {\cite{BenHamadou2022}}           & \begin{tabular}[c]{@{}c@{}}92.63 \\ (92.18, 93.1)\end{tabular}          & \begin{tabular}[c]{@{}c@{}}91.39 \\ (90.70, 92.07)\end{tabular}            & \begin{tabular}[c]{@{}c@{}}91.87 \\ (91.19, 92.56)\end{tabular}             &\begin{tabular}[c]{@{}c@{}}91.87 \\ (90.89, 92.85)\end{tabular}            & \begin{tabular}[c]{@{}c@{}}\textbf{94.82} \\ \textbf{(94.08, 95.57)}\end{tabular}            & \begin{tabular}[c]{@{}c@{}}90.93 \\ (89.97, 91.89)\end{tabular}          & \begin{tabular}[c]{@{}c@{}}93.79 \\ (93.19, 94.38)\end{tabular} & \begin{tabular}[c]{@{}c@{}}\textbf{92.65} \\ \textbf{(91.78, 93.52)}\end{tabular}          & \begin{tabular}[c]{@{}c@{}}\textbf{92.31} \\ \textbf{(91.69, 92.94)}\end{tabular}            \\
                                                              & PointNet   {\cite{qi2017pointnet}}        & \begin{tabular}[c]{@{}c@{}}72.97 \\ (71.31, 74.63)\end{tabular}          & \begin{tabular}[c]{@{}c@{}}75.46 \\ (73.87, 77.05)\end{tabular}            & \begin{tabular}[c]{@{}c@{}}77.41 \\ (76.12, 78.71)\end{tabular}            & \begin{tabular}[c]{@{}c@{}}76.36 \\ (74.41, 78.31)\end{tabular}            & \begin{tabular}[c]{@{}c@{}}75.20 \\ (73.10, 77.30)\end{tabular}          & \begin{tabular}[c]{@{}c@{}}70.25 \\ (67.58, 72.92)\end{tabular}                                            & \begin{tabular}[c]{@{}c@{}}67.09 \\ (64.17, 70.00)\end{tabular}          & \begin{tabular}[c]{@{}c@{}}64.91 \\ (60.74, 69.08)\end{tabular}          & \begin{tabular}[c]{@{}c@{}}62.49 \\ (57.45, 67.53)\end{tabular}            \\
                                                              & DGCNN   {\cite{wang2019dgcnn}}           & \begin{tabular}[c]{@{}c@{}}83.20 \\ (82.16, 84.24)\end{tabular}          & \begin{tabular}[c]{@{}c@{}}84.35 \\ (82.97, 85.73)\end{tabular}            & \begin{tabular}[c]{@{}c@{}}83.90 \\ (82.61, 85.19)\end{tabular}            & \begin{tabular}[c]{@{}c@{}}84.21 \\ (82.89, 85.54)\end{tabular}            & \begin{tabular}[c]{@{}c@{}}84.33 \\ (83.24, 85.43)\end{tabular}          & \begin{tabular}[c]{@{}c@{}}81.14 \\ (79.60, 82.68)\end{tabular}                                            & \begin{tabular}[c]{@{}c@{}}83.76 \\ (82.15, 85.37)\end{tabular}          & \begin{tabular}[c]{@{}c@{}}78.08 \\ (74.93, 81.22)\end{tabular}          & \begin{tabular}[c]{@{}c@{}}66.62 \\ (63.25, 69.88)\end{tabular}            \\
                                                              & TSegFormer   {\cite{lee2023tsegformer}}      & \begin{tabular}[c]{@{}c@{}}65.58 \\ (64.47, 66.68)\end{tabular}          & \begin{tabular}[c]{@{}c@{}}63.73  \\ (62.23, 65.22)\end{tabular}                                & \begin{tabular}[c]{@{}c@{}}62.59 \\ (60.94, 64.24)\end{tabular}            & \begin{tabular}[c]{@{}c@{}}61.53 \\ (59.77, 63.29)\end{tabular}            & \begin{tabular}[c]{@{}c@{}}70.77 \\ (69.28, 72.25)\end{tabular}          & \begin{tabular}[c]{@{}c@{}}68.42 \\ (66.61, 70.24)\end{tabular}                                            & \begin{tabular}[c]{@{}c@{}}64.41 \\ (62.22, 66.60)\end{tabular}          & \begin{tabular}[c]{@{}c@{}}62.43 \\ (59.25, 65.61)\end{tabular}          & \begin{tabular}[c]{@{}c@{}}66.11 \\ (62.61, 69.62)\end{tabular}            \\ \hline
                \multirow{5}{*}[1.5ex]{\rotatebox[origin=c]{90}{\parbox{4cm}{%
            \centering {TADPM {\cite{Lei2024AutomaticTADPM}} (Internal)}}}}     & Ours                      & \begin{tabular}[c]{@{}c@{}}\textbf  {96.49} \\ \textbf{(96.26, 96.73)}\end{tabular} & \begin{tabular}[c]{@{}c@{}}95.58 \\ (95.33, 95.83)\end{tabular}          & \begin{tabular}[c]{@{}c@{}}95.91 \\ (95.66, 96.16)\end{tabular}            & \begin{tabular}[c]{@{}c@{}}96.64 \\ (96.35, 96.93)\end{tabular}            & \begin{tabular}[c]{@{}c@{}}\textbf{96.32} \\ \textbf{(95.99, 96.66)}\end{tabular} & \begin{tabular}[c]{@{}c@{}}\textbf{96.11} \\ \textbf{(95.79, 96.43)}\end{tabular}                                   & \begin{tabular}[c]{@{}c@{}}\textbf{97.39} \\ \textbf{(97.13, 97.65)}\end{tabular} & \begin{tabular}[c]{@{}c@{}}\textbf{97.59} \\ \textbf{(97.25, 97.93)}\end{tabular} & \begin{tabular}[c]{@{}c@{}}91.68 \\ (90.31, 93.04)\end{tabular}            \\
                                                              & TGNet   {\cite{BenHamadou2022}}           & \begin{tabular}[c]{@{}c@{}}92.78 \\ (92.24, 93.33)\end{tabular}          & \begin{tabular}[c]{@{}c@{}}\textbf{97.21} \\ \textbf{(96.58, 97.84)}\end{tabular}   & \begin{tabular}[c]{@{}c@{}}\textbf{97.79} \\ \textbf{(97.09, 98.50)}\end{tabular}   & \begin{tabular}[c]{@{}c@{}}\textbf{98.08} \\ \textbf{(97.62, 98.55)}\end{tabular}   & \begin{tabular}[c]{@{}c@{}}96.20 \\ (95.14, 97.27)\end{tabular}          & \begin{tabular}[c]{@{}c@{}}73.27 \\ (71.60, 74.93)\end{tabular}                                            & \begin{tabular}[c]{@{}c@{}}90.39 \\ (89.58, 91.20)\end{tabular}          & \begin{tabular}[c]{@{}c@{}}96.06 \\ (95.24, 96.88)\end{tabular}          & \begin{tabular}[c]{@{}c@{}}\textbf{92.76}  \\ \textbf{(91.22, 94.29)}\end{tabular}\\
                                                              & PointNet   {\cite{qi2017pointnet}}        & \begin{tabular}[c]{@{}c@{}}89.49 \\ (88.98, 90.01)\end{tabular}          & \begin{tabular}[c]{@{}c@{}}90.78 \\ (90.19, 91.37)\end{tabular}            & \begin{tabular}[c]{@{}c@{}}89.61 \\ (89.01, 90.22)\end{tabular}            & \begin{tabular}[c]{@{}c@{}}90.20 \\ (89.50, 90.90)\end{tabular}            & \begin{tabular}[c]{@{}c@{}}87.44 \\ (86.66, 88.23)\end{tabular}          & \begin{tabular}[c]{@{}c@{}}87.00 \\ (86.24, 87.76)\end{tabular}                                            & \begin{tabular}[c]{@{}c@{}}90.93 \\ (90.21, 91.64)\end{tabular}          & \begin{tabular}[c]{@{}c@{}}91.32 \\ (90.51, 92.14)\end{tabular}          & \begin{tabular}[c]{@{}c@{}}71.79 \\ (70.41, 73.17)\end{tabular}            \\
                                                              & DGCNN   {\cite{wang2019dgcnn}}           & \begin{tabular}[c]{@{}c@{}}88.33 \\ (87.32, 89.34)\end{tabular}          & \begin{tabular}[c]{@{}c@{}}91.82 \\ (90.93, 92.71)\end{tabular}            & \begin{tabular}[c]{@{}c@{}}91.32 \\ (90.38, 92.26)\end{tabular}            & \begin{tabular}[c]{@{}c@{}}91.63 \\ (90.58, 92.68)\end{tabular}            & \begin{tabular}[c]{@{}c@{}}84.20 \\ (82.44, 85.95)\end{tabular}          & \begin{tabular}[c]{@{}c@{}}84.17 \\ (82.48, 85.86)\end{tabular}                                            & \begin{tabular}[c]{@{}c@{}}87.85 \\ (86.12, 89.59)\end{tabular}          & \begin{tabular}[c]{@{}c@{}}91.78 \\ (90.74, 92.82)\end{tabular}          & \begin{tabular}[c]{@{}c@{}}37.31 \\ (34.88, 39.74)\end{tabular}            \\
                                                              & TSegFormer   {\cite{lee2023tsegformer}}      & \begin{tabular}[c]{@{}c@{}}74.30 \\ (73.84, 74.75)\end{tabular}          & \begin{tabular}[c]{@{}c@{}}75.89 \\ (75.09, 76.69)\end{tabular}            & \begin{tabular}[c]{@{}c@{}}74.44 \\ (73.70, 75.18)\end{tabular}            & \begin{tabular}[c]{@{}c@{}}74.43 \\ (73.59, 75.27)\end{tabular}            & \begin{tabular}[c]{@{}c@{}}79.21 \\ (78.36, 80.05)\end{tabular}          & \begin{tabular}[c]{@{}c@{}}73.47 \\ (72.44, 74.51)\end{tabular}                                             & \begin{tabular}[c]{@{}c@{}}76.54 \\ (75.85, 77.23)\end{tabular}          & \begin{tabular}[c]{@{}c@{}}66.70 \\ (65.43, 67.96)\end{tabular}          & \begin{tabular}[c]{@{}c@{}}60.00 \\ (57.69, 62.31)\end{tabular}            \\ \hline
                \multirow{5}{*}[1.5ex]{\rotatebox[origin=c]{90}{\parbox{4cm}{%
            \centering {3D-IOSSeg {\cite{Li20243D-IOSSeg}} (External)}}}} & Ours                      & \begin{tabular}[c]{@{}c@{}}\textbf{93.60} \\ \textbf{(93.05, 94.16)}\end{tabular} & \begin{tabular}[c]{@{}c@{}}\textbf{93.64} \\ \textbf{(93.10, 94.17)}\end{tabular}   & \begin{tabular}[c]{@{}c@{}}\textbf{93.03} \\ \textbf{(92.24, 93.82)}\end{tabular}   & \begin{tabular}[c]{@{}c@{}}\textbf{94.90} \\ \textbf{(94.58, 95.22)}\end{tabular}   & \begin{tabular}[c]{@{}c@{}}\textbf{95.26} \\ \textbf{(94.75, 95.78)}\end{tabular} & \begin{tabular}[c]{@{}c@{}}\textbf{94.70} \\ \textbf{(94.11, 95.30)}\end{tabular}                                   & \begin{tabular}[c]{@{}c@{}}\textbf{95.25} \\ \textbf{(94.52, 95.97)}\end{tabular} & \begin{tabular}[c]{@{}c@{}}\textbf{93.22} \\ \textbf{(91.91, 94.54)}\end{tabular} & \begin{tabular}[c]{@{}c@{}}55.22 \\ (48.19, 62.26)\end{tabular}            \\
                                                              & TGNet   {\cite{BenHamadou2022}}           & \begin{tabular}[c]{@{}c@{}}86.86 \\ (84.92, 88.80)\end{tabular}          & \begin{tabular}[c]{@{}c@{}}91.24 \\ (90.09, 92.38)\end{tabular}            & \begin{tabular}[c]{@{}c@{}}90.79 \\ (88.92, 92.67)\end{tabular}            & \begin{tabular}[c]{@{}c@{}}93.25 \\ (91.95, 94.55)\end{tabular}            & \begin{tabular}[c]{@{}c@{}}92.78 \\ (90.33, 95.23)\end{tabular}          & \begin{tabular}[c]{@{}c@{}}74.22 \\ (69.72, 78.72)\end{tabular}                                            & \begin{tabular}[c]{@{}c@{}}83.45 \\ (79.85, 87.05)\end{tabular}          & \begin{tabular}[c]{@{}c@{}}87.08 \\ (83.34, 90.77)\end{tabular}          & \begin{tabular}[c]{@{}c@{}}\textbf{63.24} \\ \textbf{(56.14, 70.34)}\end{tabular}   \\
                                                              & PointNet   {\cite{qi2017pointnet}}        & \begin{tabular}[c]{@{}c@{}}77.09 \\ (75.73, 78.45)\end{tabular}          & \begin{tabular}[c]{@{}c@{}}76.14 \\ (74.38, 77.91)\end{tabular}            & \begin{tabular}[c]{@{}c@{}}76.00 \\ (74.30, 77.69)\end{tabular}            & \begin{tabular}[c]{@{}c@{}}79.62 \\ (78.22, 81.03)\end{tabular}            & \begin{tabular}[c]{@{}c@{}}80.33 \\ (78.82, 81.83)\end{tabular}          & \begin{tabular}[c]{@{}c@{}}79.02 \\ (77.09, 80.94)\end{tabular}                                            & \begin{tabular}[c]{@{}c@{}}81.89 \\ (80.13, 83.66)\end{tabular}          & \begin{tabular}[c]{@{}c@{}}71.76 \\ (68.60, 74.91)\end{tabular}          & \begin{tabular}[c]{@{}c@{}}32.78 \\ (27.58, 37.98)\end{tabular}            \\
                                                              & DGCNN   {\cite{wang2019dgcnn}}           & \begin{tabular}[c]{@{}c@{}}83.88 \\ (82.66, 85.11)\end{tabular}          & \begin{tabular}[c]{@{}c@{}}84.40 \\ (82.87, 85.94)\end{tabular}            & \begin{tabular}[c]{@{}c@{}}83.76 \\ (82.14, 85.39)\end{tabular}            & \begin{tabular}[c]{@{}c@{}}84.03 \\ (82.53, 85.54)\end{tabular}            & \begin{tabular}[c]{@{}c@{}}83.74 \\ (82.14, 85.33)\end{tabular}          & \begin{tabular}[c]{@{}c@{}}84.18 \\ (82.59, 85.77)\end{tabular}                                            & \begin{tabular}[c]{@{}c@{}}87.63 \\ (86.15, 89.12)\end{tabular}          & \begin{tabular}[c]{@{}c@{}}85.20 \\ (83.61, 86.80)\end{tabular}          & \begin{tabular}[c]{@{}c@{}}38.14 \\ (33.36, 42.93)\end{tabular}            \\
                                                              & TSegFormer   {\cite{lee2023tsegformer}}      & \begin{tabular}[c]{@{}c@{}}62.04 \\ (61.06, 63.01)\end{tabular}          & \begin{tabular}[c]{@{}c@{}}57.01 \\ (55.66, 58.35)\end{tabular}            & \begin{tabular}[c]{@{}c@{}}56.49 \\ (54.85, 58.13)\end{tabular}            & \begin{tabular}[c]{@{}c@{}}62.13 \\ (60.71, 63.55)\end{tabular}            & \begin{tabular}[c]{@{}c@{}}70.16 \\ (68.49, 71.84)\end{tabular}          & \begin{tabular}[c]{@{}c@{}}67.15 \\ (64.97, 69.33)\end{tabular}                                            & \begin{tabular}[c]{@{}c@{}}62.89 \\ (60.73, 65.04)\end{tabular}          & \begin{tabular}[c]{@{}c@{}}61.76 \\ (59.64, 63.88)\end{tabular}          & \begin{tabular}[c]{@{}c@{}}41.09 \\ (35.49, 46.68)\end{tabular}            
    \end{tabular}
    }
\end{sidewaystable*}
\clearpage
\begin{sidewaystable*}[h]
    \centering
    \caption{Ablation results comparing pretrained initialization and training from scratch. Dice score (\%) is reported as mean (95\% confidence interval). Bold values indicate the best performance.}
    \label{table:ablation_results}
    \resizebox{\textwidth}{!}{%
    \begin{tabular}{c|c|c|c|c|c|c|c|c|c|c}
        \multirow{2}{*}{Dataset}                             & \multirow{2}{*}{Method} & \multirow{2}{*}{Total ↑}      & T1 ↑                            & T2 ↑                            & T3 ↑                            & T4 ↑                          & T5 ↑                                                            & T6 ↑                          & T7 ↑                          & T8 ↑                            \\
                                                      &                           &                               & (Central Incisor)               & (Lateral   Incisor)             & (Cuspid)                        & (1st   Premolar)              & (2nd   Premolar)                                                & (1st   Molar)                 & (2nd   Molar)                 & (3rd   Molar)                   \\ \hline
        \multirow{2}{*}{ToothFairy2 \cite{2024TMI}} & Ours                 & \begin{tabular}[c]{@{}c@{}}\textbf{93.38} \\ \textbf{(92.02, 94.73)}\end{tabular} & \begin{tabular}[c]{@{}c@{}}\textbf{96.21} \\ \textbf{(95.34, 97.08)}\end{tabular} & \begin{tabular}[c]{@{}c@{}}\textbf{96.10} \\ \textbf{(95.03, 97.17)}\end{tabular} & \begin{tabular}[c]{@{}c@{}}\textbf{94.37} \\ \textbf{(92.01, 96.74)}\end{tabular} & \begin{tabular}[c]{@{}c@{}}\textbf{92.33} \\ \textbf{(90.43, 94.23)}\end{tabular} & \begin{tabular}[c]{@{}c@{}}\textbf{92.83} \\ \textbf{(90.90, 94.76)}\end{tabular}          & \begin{tabular}[c]{@{}c@{}}\textbf{92.14} \\ \textbf{(89.93, 94.36)}\end{tabular}   & \begin{tabular}[c]{@{}c@{}}\textbf{91.56} \\ \textbf{(89.31, 93.80)}\end{tabular}          & \begin{tabular}[c]{@{}c@{}}\textbf{92.92} \\ \textbf{(90.28, 95.55)}\end{tabular}          \\
                                    & Ours w/o pretraining & \begin{tabular}[c]{@{}c@{}}88.28 \\ (86.50, 90.06)\end{tabular}          & \begin{tabular}[c]{@{}c@{}}86.96 \\ (84.51, 89.41)\end{tabular}          & \begin{tabular}[c]{@{}c@{}}90.13 \\ (88.35, 91.91)\end{tabular}          & \begin{tabular}[c]{@{}c@{}}91.65 \\ (89.07, 94.23)\end{tabular}          & \begin{tabular}[c]{@{}c@{}}88.37 \\ (85.69, 91.05)\end{tabular}          & \begin{tabular}[c]{@{}c@{}}88.21 \\ (85.36, 91.05)\end{tabular}          & \begin{tabular}[c]{@{}c@{}}83.04 \\ (78.47, 87.62)\end{tabular}          & \begin{tabular}[c]{@{}c@{}}85.75 \\ (82.47, 89.02)\end{tabular}          & \begin{tabular}[c]{@{}c@{}}88.68 \\ (84.95, 92.41)\end{tabular}          \\ \hline
        \multirow{2}{*}{Cui \cite{Cui2022FullyAutomaticAI}}  & Ours                 & \begin{tabular}[c]{@{}c@{}}\textbf{88.92} \\ \textbf{(87.97, 89.87)}\end{tabular} & \begin{tabular}[c]{@{}c@{}}\textbf{88.10} \\ \textbf{(86.30, 89.90)}\end{tabular} & \begin{tabular}[c]{@{}c@{}}\textbf{89.25} \\ \textbf{(88.14, 90.36)}\end{tabular} & \begin{tabular}[c]{@{}c@{}}\textbf{91.32} \\ \textbf{(89.84, 92.79)}\end{tabular} & \begin{tabular}[c]{@{}c@{}}\textbf{89.89} \\ \textbf{(88.77, 91.00)}\end{tabular} & \begin{tabular}[c]{@{}c@{}}\textbf{87.67} \\ \textbf{(86.12, 89.22)}\end{tabular} & \begin{tabular}[c]{@{}c@{}}\textbf{90.81} \\ \textbf{(90.09, 91.54)}\end{tabular} & \begin{tabular}[c]{@{}c@{}}\textbf{90.39} \\ \textbf{(89.39, 91.39)}\end{tabular} & \begin{tabular}[c]{@{}c@{}}\textbf{88.85} \\ \textbf{(87.27, 90.42)}\end{tabular} \\
                                    & Ours w/o pretraining & \begin{tabular}[c]{@{}c@{}}85.83 \\ (84.43, 87.23)\end{tabular}          & \begin{tabular}[c]{@{}c@{}}87.01 \\ (85.02, 89.00)\end{tabular}          & \begin{tabular}[c]{@{}c@{}}87.80 \\ (86.23, 89.36)\end{tabular}          & \begin{tabular}[c]{@{}c@{}}90.94 \\ (89.88, 91.99)\end{tabular}          & \begin{tabular}[c]{@{}c@{}}86.67 \\ (84.25, 89.09)\end{tabular}          & \begin{tabular}[c]{@{}c@{}}83.35 \\ (81.06, 85.65)\end{tabular}          & \begin{tabular}[c]{@{}c@{}}85.97 \\ (83.74, 88.19)\end{tabular}          & \begin{tabular}[c]{@{}c@{}}84.33 \\ (82.05, 86.60)\end{tabular}          & \begin{tabular}[c]{@{}c@{}}82.55 \\ (80.06, 85.05)\end{tabular}          \\ \hline
        \multirow{2}{*}{Teeth3DS \cite{BenHamadou2022}}    & Ours                 & \begin{tabular}[c]{@{}c@{}}\textbf{93.16} \\ \textbf{(92.58, 93.74)}\end{tabular} & \begin{tabular}[c]{@{}c@{}}\textbf{93.61} \\ \textbf{(92.79, 94.44)}\end{tabular}   & \begin{tabular}[c]{@{}c@{}}\textbf{93.42} \\ \textbf{(92.67, 94.17)}\end{tabular} & \begin{tabular}[c]{@{}c@{}}\textbf{92.71} \\ \textbf{(91.68, 93.73)}\end{tabular} & \begin{tabular}[c]{@{}c@{}}\textbf{92.70} \\ \textbf{(91.75, 93.65)}\end{tabular}            & \begin{tabular}[c]{@{}c@{}}\textbf{92.63} \\ \textbf{(91.72, 93.54)}\end{tabular} & \begin{tabular}[c]{@{}c@{}}\textbf{94.22} \\ \textbf{(93.44, 95.00)}\end{tabular} & \begin{tabular}[c]{@{}c@{}}\textbf{91.20} \\ \textbf{(89.85, 92.55)}\end{tabular} & \begin{tabular}[c]{@{}c@{}}\textbf{84.65} \\ \textbf{(82.82, 86.49)}\end{tabular} \\
                                    & Ours w/o pretraining & \begin{tabular}[c]{@{}c@{}}87.86 \\ (85.77, 89.94)\end{tabular}          & \begin{tabular}[c]{@{}c@{}}91.72 \\ (89.67, 93.76)\end{tabular}          & \begin{tabular}[c]{@{}c@{}}89.77 \\ (87.57, 91.98)\end{tabular}          & \begin{tabular}[c]{@{}c@{}}89.25 \\ (87.01, 91.48)\end{tabular}          & \begin{tabular}[c]{@{}c@{}}88.49 \\ (85.98, 91.00)\end{tabular}          & \begin{tabular}[c]{@{}c@{}}83.89 \\ (80.52, 87.27)\end{tabular}          & \begin{tabular}[c]{@{}c@{}}86.39 \\ (83.22, 89.56)\end{tabular}          & \begin{tabular}[c]{@{}c@{}}80.92 \\ (76.80, 85.04)\end{tabular}          & \begin{tabular}[c]{@{}c@{}}67.44 \\ (61.64, 73.25)\end{tabular}          \\ \hline
        \multirow{2}{*}{TADPM \cite{Lei2024AutomaticTADPM}}       & Ours                 & \begin{tabular}[c]{@{}c@{}}\textbf{96.49} \\ \textbf{(96.26, 96.73)}\end{tabular} & \begin{tabular}[c]{@{}c@{}}\textbf{95.58} \\ \textbf{(95.33, 95.83)}\end{tabular}          & \begin{tabular}[c]{@{}c@{}}\textbf{95.91} \\ \textbf{(95.66, 96.16)}\end{tabular}          & \begin{tabular}[c]{@{}c@{}}96.64 \\ (96.35, 96.93)\end{tabular}          & \begin{tabular}[c]{@{}c@{}}\textbf{96.32} \\ \textbf{(95.99, 96.66)}\end{tabular} & \begin{tabular}[c]{@{}c@{}}\textbf{96.11} \\ \textbf{(95.79, 96.43)}\end{tabular} & \begin{tabular}[c]{@{}c@{}}\textbf{97.39} \\ \textbf{(97.13, 97.65)}\end{tabular} & \begin{tabular}[c]{@{}c@{}}\textbf{97.59} \\ \textbf{(97.25, 97.93)}\end{tabular} & \begin{tabular}[c]{@{}c@{}}\textbf{91.68} \\ \textbf{(90.31, 93.04)}\end{tabular}          \\
                                    & Ours w/o pretraining & \begin{tabular}[c]{@{}c@{}}95.85 \\ (95.49, 96.22)\end{tabular}          & \begin{tabular}[c]{@{}c@{}}94.98 \\ (94.38, 95.57)\end{tabular}          & \begin{tabular}[c]{@{}c@{}}95.24 \\ (94.67, 95.81)\end{tabular}          & \begin{tabular}[c]{@{}c@{}}\textbf{96.86} \\ \textbf{(96.63, 97.08)}\end{tabular}          & \begin{tabular}[c]{@{}c@{}}94.52 \\ (93.40, 95.64)\end{tabular}          & \begin{tabular}[c]{@{}c@{}}95.31 \\ (94.63, 96.00)\end{tabular}          & \begin{tabular}[c]{@{}c@{}}96.95 \\ (96.51, 97.40)\end{tabular}          & \begin{tabular}[c]{@{}c@{}}97.01 \\ (96.40, 97.62)\end{tabular}          & \begin{tabular}[c]{@{}c@{}}88.91 \\ (86.66, 91.15)\end{tabular}          \\ \hline
        \multirow{2}{*}{3D-IOSSeg \cite{Li20243D-IOSSeg}}   & Ours                 & \begin{tabular}[c]{@{}c@{}}\textbf{93.60} \\ \textbf{(93.05, 94.16)}\end{tabular} & \begin{tabular}[c]{@{}c@{}}\textbf{93.64} \\ \textbf{(93.10, 94.17)}\end{tabular} & \begin{tabular}[c]{@{}c@{}}\textbf{93.03} \\ \textbf{(92.24, 93.82)}\end{tabular} & \begin{tabular}[c]{@{}c@{}}\textbf{94.90} \\ \textbf{(94.58, 95.22)}\end{tabular} & \begin{tabular}[c]{@{}c@{}}\textbf{95.27} \\ \textbf{(94.75, 95.78)}\end{tabular} & \begin{tabular}[c]{@{}c@{}}\textbf{94.70} \\ \textbf{(94.11, 95.30)}\end{tabular} & \begin{tabular}[c]{@{}c@{}}\textbf{95.25} \\ \textbf{(94.52, 95.97)}\end{tabular} & \begin{tabular}[c]{@{}c@{}}\textbf{93.22} \\ \textbf{(91.91, 94.54)}\end{tabular} & \begin{tabular}[c]{@{}c@{}}\textbf{55.22} \\ \textbf{(48.19, 62.26)}\end{tabular}          \\
                                    & Ours w/o pretraining & \begin{tabular}[c]{@{}c@{}}90.01 \\ (88.11, 91.91)\end{tabular}          & \begin{tabular}[c]{@{}c@{}}92.09 \\ (90.97, 93.21)\end{tabular}            & \begin{tabular}[c]{@{}c@{}}91.09 \\ (89.38, 92.79)\end{tabular}          & \begin{tabular}[c]{@{}c@{}}93.29 \\ (91.89, 94.69)\end{tabular}          & \begin{tabular}[c]{@{}c@{}}91.43 \\ (88.37, 94.48)\end{tabular}          & \begin{tabular}[c]{@{}c@{}}88.97 \\ (85.30, 92.64)\end{tabular}          & \begin{tabular}[c]{@{}c@{}}89.76 \\ (86.62, 92.91)\end{tabular}          & \begin{tabular}[c]{@{}c@{}}89.70 \\ (87.13, 92.28)\end{tabular}          & \begin{tabular}[c]{@{}c@{}}52.78 \\ (46.21, 59.34)\end{tabular}         
    \end{tabular}
    }
\end{sidewaystable*}
\clearpage
\end{appendices}
\end{document}